%% file: arxiv.tex
\documentclass[10pt,twocolumn,letterpaper]{article}

\usepackage[final]{arxiv}      

\input{preamble}

\usepackage{lipsum}
\usepackage{comment}
\usepackage{multirow}
\usepackage{caption}
\usepackage{rotating}
\usepackage{scalerel}
\include{symbols_format}

\usepackage[pagebackref,breaklinks,colorlinks]{hyperref}

\title{Developability Approximation for Neural Implicits through Rank Minimization}

 \author{Pratheba Selvaraju\\
 University of Massachusetts, Amherst\\
 {\tt\small pselvaraju@cs.umass.edu}
 }

\begin{document}

\twocolumn[{%
\renewcommand\twocolumn[1][]{#1}%
\maketitle
\begin{center}
    \centering
    \captionsetup{type=figure}
    \includegraphics[width=1\linewidth]{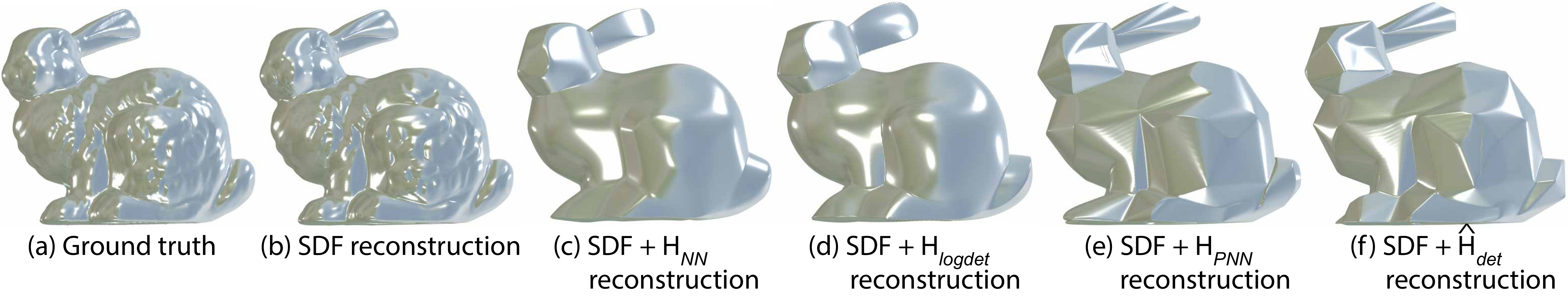}
    \caption{\label{fig:teaser}{\textbf{Reconstructed developable approximation of implicit surface}. We propose Implicit developable surface approximation using rank minimization heuristics as regularizer on second order differentiable of the objective function. We are able to achieve approximate piecewise developable surfaces with automatic crease emergence. The figure.\ref{fig:teaser} shows (a) the ground truth surface, (b) implicit (signed distance field - SDF) reconstruction without regularizer, and (c) - (f) implicit reconstruction with regularizers detailed in \ref{para:reg_term}.}}
\end{center}%
}]
\maketitle
\input{sec/0_abstract}    
\input{sec/1_intro}

\input{sec/2_related}

\input{sec/3_background}

\input{sec/4_method}

\input{sec/5_experiments}

\input{sec/7_conclusion}

{
    \small
    \bibliographystyle{ieeenat_fullname}
    \bibliography{main}
}
\input{sec/X_suppl}

\end{document}

%% file: preamble.tex
\usepackage[dvipsnames]{xcolor}


%% file: symbols_format.tex
\usepackage[dvipsnames]{xcolor}

\newcommand{\bn}{\mathbf{n}}

\newcommand{\bp}{\mathbf{p}}

\newcommand{\bA}{\mathbf{A}}
\newcommand{\bB}{\mathbf{B}}

\newcommand{\bH}{\mathbf{H}}
\newcommand{\bI}{\mathbf{I}}

\newcommand{\bX}{\mathbf{X}}

\newcommand{\btheta}{\mbox{\boldmath$\theta$}}

\newcommand{\mR}{\mathcal{R}}

\newcommand{\mP}{\mathcal{P}}

\definecolor{VangelisColor}{rgb}{0,0,0.8} 

\definecolor{PrathebaColor}{rgb}{1.0,0,0} 

%% file: sec/0_abstract.tex
\begin{abstract} 
\vspace{-2mm}
Developability refers to the process of creating a surface without any tearing or shearing from a two-dimensional plane. It finds practical applications in the fabrication industry. An essential characteristic of a developable 3D surface is its zero Gaussian curvature, which means that either one or both of the principal curvatures are zero. This paper introduces a method for reconstructing an approximate developable surface from a neural implicit surface. The central idea of our method involves incorporating a regularization term that operates on the second-order derivatives of the neural implicits, effectively promoting zero Gaussian curvature. Implicit surfaces offer the advantage of smoother deformation with infinite resolution, overcoming the high polygonal constraints of state-of-the-art methods using discrete representations. We draw inspiration from the properties of surface curvature and employ rank minimization techniques derived from compressed sensing. Experimental results on both developable and non-developable surfaces, including those affected by noise, validate the generalizability of our method.
\vspace{-2mm}
\end{abstract}

%% file: sec/1_intro.tex
\section{Introduction}
\label{sec:intro}
Developable surfaces have practical applications in digital fabrication, industrial design, architecture, and geometric surface abstraction. They are used in the automotive industry for car body panels, in industrial design for curved furniture elements, in computer graphics to simplify complex 3D shapes to enhance rendering performance, and in architecture for building facades to reduce material wastage. Research efforts are focused on developing computer-aided algorithms to identify these developable patches on a surface, aiming to minimize material wastage in industry applications during cutting and reassembling processes. Several approaches exist for approximating developable shapes, employing optimization techniques \cite{Odedstein:DevelopabilityTriangleMeshes:2018, Sellan:Developability:2020, BinningerVerhoeven:GaussThinning:2021, Wang:Achievingdevelopability:2004}, shape wrapping techniques that utilize multiple developable patches \cite{Ion:ApproximatingDOGs:2020, Mitani:MakingPapercraft:2004, Shatz:PaperCraftModels:2006} and methods for synthesizing developable surfaces \cite{Tang:InteractiveDesign:2016}. Existing methods for reconstructing developable shapes often rely on fixed topology representations. In contrast, implicit surfaces offer advantages such as smooth interpolation, deformation capabilities, and the ability to handle topological changes naturally. In recent years, there has been extensive research in neural implicit surface representations for 3D reconstruction, from single shape reconstruction \cite{Carr:ReconRadial:2001}, to data-driven methods \cite{Park:deepsdf:2019, sitzmann:siren:2019}, and generative modeling by deformation \cite{Chen:LearningImplicitFields:2019}. These approaches achieve high-detail surface reconstruction, and regularization techniques have been incorporated to encourage smoothness \cite{Gropp:ImplicitGeometricRegularization:2020}. However, there is currently no implicit reconstruction method that specifically promotes surface developability. Our paper presents a novel approach where we introduce a regularization term into neural implicit formulations to deform the surface into approximately piecewise developable patches.

Our approach is driven by two key observations. Firstly, implicit surfaces offer the advantage of providing access to gradients and higher-order derivatives, allowing us to compute surface normals, curvature, and other surface properties without additional computations. In our method, we utilize the second-order derivatives of the implicit function with respect to input coordinates, which are related to Gaussian curvature \cite{Goldman:CurvatureImplicit:2005}. Secondly, surface developability necessitates zero Gaussian curvature \cite{docarmo:DiffGeom:1976}. To achieve this, we formulate the developability condition for implicits as a rank minimization problem, inspired by the work of Sellan et al. \cite{Sellan:Developability:2020}, who applied the concept to height fields derived from depth maps. We employ both gaussian curvature minimization and implicit hessian rank minimization, combining them in an objective that encourages both developability, resulting in the elimination of Gaussian curvature and shape fitting of the implicit to the input point cloud.

Our implicit-driven approach has key advantages over current discrete representation methods, which struggle with higher polygon counts. Discrete optimization-based methods suffer from deformation-induced topology changes leading to inaccuracies surface representation, our approach allows for easier topological deformation while maintaining shape approximation. Additionally, our method eliminates the need for specialized solvers by relying on a single regularizer weight value to control the level of developability.

In summary, our main contribution is a novel approach introducing regularization term and optimization procedure promoting developability in neural implicit surface reconstruction. Qualitative and quantitative evaluations in section \ref{para:results} demonstrate its effectiveness for complex topologies, robustness to noise, and superior shape preservation compared to existing methods.

%% file: sec/2_related.tex
\section{Related Works}
Our approach leverages implicit functions and employs second-order optimization techniques based on rank minimization to reconstruct an input point cloud into a piecewise developable surface. In this section, we provide a brief overview of prior research on developable surface reconstructions, and applications of rank minimization approximations.
\vspace{-2mm}
\paragraph{Developable surface Approximation.}
Existing developable surface reconstruction methods often use discrete representations and rely on constrained optimization or patch-based wrapping. For instance, Stein et al. \cite{Odedstein:DevelopabilityTriangleMeshes:2018} achieve piece-wise developable surfaces through mesh vertex optimization. However, these methods assume noise-free inputs and may require manual tuning, potentially getting trapped in local minima due to non-convexity. Binninger and Verhoeven et al. \cite{BinningerVerhoeven:GaussThinning:2021} use Gaussian image thinning for developability with automatic crease development. However, increasing cone angles leads to significant shape deformation, deviating from the ground truth shape. Gavriil et al. \cite{Gavriil:OptimizingBSpline:2019} propose a similar approach using Gauss thinning. Sellen et al. \cite{Sellan:Developability:2020} achieve developability with nuclear norm minimization but focus on planar height fields. Computation time in these methods scales with polygon count in the mesh.
Patch wrapping methods approximate a surface by fitting developable patches onto the input surface geometry. Verhoeven et al. \cite{Verhoeven:Dev2PQ:2022} use planar quad strips aligned to principal curvatures for curved parts, while Rabinovich et al. \cite{Rabinovich:DiscGeoNets:2018} use orthogonal quad meshes to optimize for developable surfaces with constrained overlapping quad patches. Peternell et al. \cite{Peternell:DevSurfToPointCloud:2004} fit a developable surface by estimating tangent planes through curve approximation from data points.
Spline-based methods offer another discrete representation for smooth reconstruction. Tang et al. \cite{Tang:InteractiveDesign:2016} use cubic spline developable patches projected onto the surface and merge them iteratively with proximity constraints. Leopoldseder et al. \cite{Leopoldseder:ApproximationOD:1998} employ interpolation of tangent planes with right circular cones for appropriate rulings. Gavriil et al. \cite{Gavriil:OptimizingBSpline:2019} propose B-spline surface optimization using Gauss thinning for developability. Kenneth et al. \cite{Kenneth:DevSurfSketchBound:2007} define a 3D polyline boundary and generate a smooth discrete developable surface that interpolates this boundary. Additionally, Solomon et al. \cite{Solomon:FlexibleDevSurf:2012} introduce a flexible structure for developable surface representation.

On the other hand, our proposed method uses implicit functions and second-order optimization based on rank minimization to achieve piecewise developable surfaces from point cloud inputs.

\paragraph{Rank minimization.}\label{para:sec2_rank_min}
Stein et al. \cite{Stein:NaturalBoundaryCond:2018} proposes minimizing the $L_1$ norm of second derivatives to reconstruct piecewise planar surfaces. Sellan et al. \cite{Sellan:Developability:2020} utilizes nuclear norm minimization for reconstructing piecewise developable surfaces from input depth maps, but their method is restricted to heightfields. Liu et al. \cite{Liu:CubicStyle:2019} employ rank minimization using the $L_1$ norm to reconstruct surfaces in a cube-like style while preserving the original shape's content. Besides $L_1$ norm minimization, log-determinant minimization is utilized as rank approximation for distance matrices \cite{Fazel:LogdetHeuristic:2003}, subspace clustering in \cite{Peng:Subspacecluster:2015}. $L_0$ minimization is employed for basis selection in \cite{Wipf:L0min:2004} and for image editing in \cite{Xu:Imagesmoothl0:2011}. Additionally, Oh et al. \cite{Oh:Partialnuclearnorm:2016} uses partial sum nuclear norm minimization for PCA.

%% file: sec/3_background.tex
\section {Background}
\label{sec:background}
\paragraph{Implicit surface representation.} In an implicit surface representation, a surface is defined as the zero level set of the implicit function. The implicit function takes as input the coordinates of a point $\bp\in \mR^3$ in space and returns a scalar value $s=f(\bp)$ where $s \in \mR$. Iso-levels of the implicit function represent surfaces in $\mR^3$. Points on the surface are those where the implicit function evaluates to zero $f(\bp)=0$. Points with negative scalar values $f(\bp)<0$ correspond to the shape interior, while points with positive values $f(\bp)>0$ represent the shape exterior.

\paragraph{Gaussian curvature and developability.} 
A developable surface is defined by its Gaussian curvature, which plays a key role in its characterization. Gaussian curvature measures the curvature of a surface at each point and is determined by the \emph{principal curvatures} \cite{docarmo:DiffGeom:1976}. In the case of a developable surface, the Gaussian curvature is zero everywhere. This means that either or both the principal curvatures, $K_1$ and $K_2$, is zero. As a result, developable surfaces can be unrolled or flattened onto a plane without any distortion or stretching.
The principal curvatures of a surface represent the extreme curvatures in different directions. They are determined by the maximum and minimum values of the \emph{normal curvature} at a given point on the surface. Normal curvature is the curvature of curves formed by the intersection of normal planes and the surface at that point. The direction with the maximum principal curvature ($K_1$) corresponds to the maximum normal curvature, while the direction with the minimum principal curvature ($K_2$) corresponds to the minimum normal curvature. For a deeper understanding of the differential geometry of surfaces and its relation to geometry processing applications, refer to the tutorial by Crane et al. \cite{Crane:DGP:2013}.

\paragraph{Gaussian curvature of implicits.} 
The Gaussian curvature, denoted as $K$, of a surface represented by implicit function $f(\bp)$ can be defined by the first and second-order derivatives of the implicit function with respect to the surface coordinates $\bp\in \mR^3$. The first order derivative, known as the gradient \mbox{$\nabla f(\bp) = ( \frac{\partial f(\bp)}{\partial x},\frac{\partial f(\bp)}{\partial y},\frac{\partial f(\bp)}{\partial z})$}, yields a 3D vector pointing to the direction of the surface normal. The second order derivative, referred to as the Hessian $H_f(\bold{p})$ provides information about the rate of change of the surface normal in different directions, which in turn determines the curvature.  
\begin{equation}
    \bH_f(\bp) =     
\begin{bmatrix} 
\frac{\partial^2 f(\bold{p})}{\partial^2 x} & 
\frac{\partial^2 f(\bold{p})}{\partial x \partial y} & 
\frac{\partial^2 f(\bold{p})}{\partial x \partial z} \\
\frac{\partial^2 f(\bold{p})}{\partial y \partial x} & 
\frac{\partial^2 f(\bold{p})}{\partial^2 y} & 
\frac{\partial^2 f(\bp)}{\partial y \partial z} \\
\frac{\partial^2 f(\bp)}{\partial z \partial x} & 
\frac{\partial^2 f(\bp)}{\partial z \partial y} & 
\frac{\partial^2 f(\bp)}{\partial^2 z} \\
 \end{bmatrix} 
\end{equation}
 
\noindent To compute the Gaussian curvature at a specific point $\mathbf{p}$ we employ the following procedure as outlined in the work by Goldman \cite{Goldman:CurvatureImplicit:2005}:
\vspace{-1mm}
\begin{equation}
K(\bold{p})\!\!=\!\!
-\frac{
det(\hat\bH_f(\bp))
}
{
\left| {\nabla f(\bp)} \right|^4
} 
,\mbox{where}\,\,
\hat \bH_f(\bp) \!\!=\!\! 
\begin{bmatrix} 
\bH_f(\bp) \!&\! \nabla {f(\bp)^T} \\
\nabla f(\bp) \!&\! \mathbf{0}
\end{bmatrix},
\label{eq:Imp_gauss_curv}
\end{equation}

\noindent For a smooth surface, zero Gaussian curvature at a point $\bp$ entails that $K(\bp) = 0 \Leftrightarrow K_1\cdot K_2 = 0 \Leftrightarrow K_1 = 0 \text{ or } K_2 = 0$.  Since the Gaussian surface is defined only on points where the gradient $f(\mathbf{p})$ is non-zero, having zero Gaussian curvature means that the determinant of the $4 \times 4$ matrix in the numerator of Eq.\ref{eq:Imp_gauss_curv} must be zero: $\text{det}(\hat{\mathbf{H}}_f(\mathbf{p})) = 0$. The equation for the determinant of the matrix $\hat{\mathbf{H}}_f(\mathbf{p})$ can be expressed as follows \cite{Goldman:CurvatureImplicit:2005}:
\begin{equation}
det(\hat{\mathbf{H}}_f(\mathbf{p})) = -\nabla f(\mathbf{p}) \cdot \text{Cof}(H_f(\mathbf{p})) \cdot \nabla f(\mathbf{p})^T
\label{eq:dethath}
\end{equation}

\noindent By utilizing the properties of the cofactor matrix, $\text{Cof}(\textbf{H}_f(\mathbf{p}))^T = det(\textbf{H}_f(\mathbf{p}))\textbf{H}_f(\mathbf{p})^{-1}$, we can  minimize the rank of $\mathbf{H}_f(\mathbf{p})$ than $\hat{\mathbf{H}}_f(\mathbf{p})$.

\paragraph{Norm minimization.} \label{para:rank_min}
Equating the determinant of an $n \times n$ matrix $\mathbf{X}$ to zero is essentially the same as ensuring that the matrix is not full rank, i.e., $rank(\mathbf{X}) < n$. Considering that the absolute value of the determinant of a square matrix is equivalent to the product of its singular values, i.e., $|det(A)| = \prod_{i=1}^{n} \sigma_i $, the rank minimization can be framed as a problem of minimizing the $L_0$ norm of the singular values. To prevent trivial solutions, the minimization process is subject to additional constraints, often expressed as a linear system $\bA \bX= \bB$
\cite{Fazel:ARM:2001,Sellan:Developability:2020}:
\begin{equation}
\min\limits_{\bA\bX=\bB}  \, rank(\bX)     \Leftrightarrow  \min\limits_{\bA\bX=\bB}  \, \lVert  \sigma (\bX)  \rVert_0
\label{eq:rank_l0}
\end{equation}
,where the vector $\sigma(\mathbf{X})$ stores the singular values of the matrix $\mathbf{X}$. The $L_0$ minimization problem is non-convex, non-differentiable, and generally non-tractable (NP-hard) \cite{Natarajan:SparseApprox:1995}. However, when dealing with the hessian rank of the implicits with rank of 3, we can make use of a smoother approximation of the $L_0$ minimization objective similar to \cite{Louizos:l0learning:2018, Bengio:EstimatingOP:2013}. Instead of employing the $L_0$ cardinality approximation, we consider a relaxed alternative in the form of the nuclear norm (also known as the $L_1$ norm) of the singular values which has been demonstrated as a tight convex approximation to the rank function \cite{Fazel:Thesis:2002}. The nuclear norm $\lVert \bX  \rVert_*$ of the matrix:
\begin{equation}
\min\limits_{\bA\bX=\bB}  \, \lVert  \sigma(\bX)  \rVert_1     \Leftrightarrow  \min\limits_{\bA\bX=\bB}  \, \lVert \bX  \rVert_*
\label{eq:rank_l1}
\end{equation}

\paragraph{Alternative approximations.} 
There is a significant drawback in solving for nuclear norm minimization. While minimizing the $L_1$ norm provides an approximation to the rank minimization problem and leads to a low-rank matrix $X$, it also simultaneously diminishes the high-variance information that includes important details like the structure of the object. Thus we also explored the use non-convex surrogate partial sum minimization method \cite{Oh:PartialSum:2016}, minimizing the sum of smaller singular values:
\begin{equation}
\min\limits_{\bA\bX=\bB} \, rank(\mathbf{X}) = \sum_{i=r+1}^{n} \sigma_i(\bX)
\label{eq:rank_pnn}
\end{equation}
,where  $\sigma_i(\mathbf{X})$ refers to the $i^{th}$ singular value, arranged in decreasing order, and the parameter $r$ determines the number of largest singular values to be excluded during the minimization process.
 To mitigate the impact of large singular values, an alternative non-convex surrogate involving the log determinant function was also employed \cite{Peng:Subspacecluster:2015}. Since the hessian matrix $\bH_f(\mathbf{p})$ of the implicit function may not always be positive semi-definite, we express the rank minimization in the form:
\vspace{-2mm}
\begin{equation}
\log(\det(\mathbf{X}^T\mathbf{X} + \mathbf{I})) = \sum_{i=1}^{n} \log(1+\sigma_i^2)
\end{equation}

\noindent Some other alternative relaxation methods include Weighted Nuclear Norm Minimization, Capped $L_1$ Norm, Schatten-p Norm, Truncated Rank Minimization, which we could not cover in our experiments. We recommend \cite{sagan:Lowrankfactor:2021} for brief review of these approaches.

%% file: sec/4_method.tex
\section {Method}
\label{sec:method}

\paragraph*{Overview.} 
Our approach takes a point cloud {$\mP=\{\bp_i, \bn_i\}_{i=1}^N$}, as input, where $\mathbf{p}_i \in \mathbb{R}^3$ represents the 3D position of a point, $\mathbf{n}_i \in \mathbb{R}^3$ is its corresponding normal, and $N$ is the total number of points. The output of our method is an implicit function $f(\mathbf{p})$, which assigns a scalar value to input points $\mathbf{p} \in \mathbb{R}^3$. The reconstructed surface is obtained by extracting the zero iso-level ($s=0$) of the implicit function using the marching cubes algorithm \cite{Lorensen:marchingcubes:1987}. Our goal is to obtain an implicit function that approximates the input point cloud and generates a surface that maximizes its developability. This implies that the resulting surface points should ideally possess negligible or zero Gaussian curvature while still retaining the overall shape of the point cloud.

\begin{figure}[t!]
  \centering
  \includegraphics[width=1.0\linewidth]{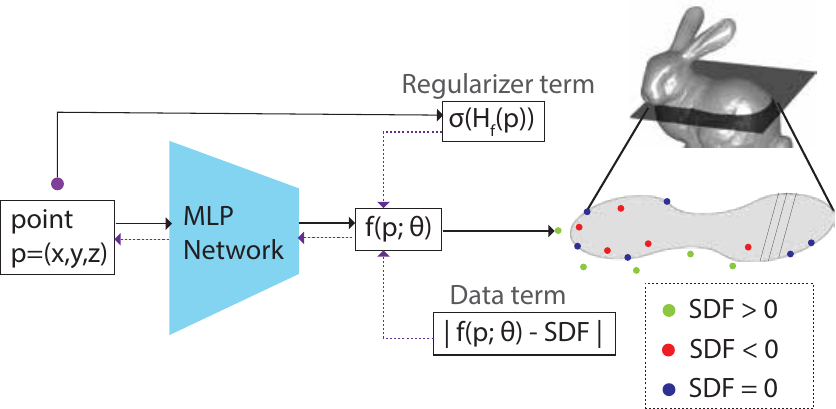}
  \vspace{-4mm}
  \caption{\label{fig:Architecture}
  \textbf{Our MLP architecture} maps a point $\bp \in \mR^3$ to an implicit function $f(\bp; \theta)$ with learned parameters $\theta$. The parameters are optimized through a loss function comprising data term and regularization term detailed in Section.\ref{sec:method}. Backpropagation (dotted arrows), computes gradients and Hessians with respect to $\bp$ coordinates for fine-tuning $\theta$.}
   \vspace{-3mm}
\end{figure}

To achieve this objective, we formulate the problem as an optimization task aimed at estimating the parameters $\boldsymbol{\theta}$ of a neural network function $f(\mathbf{p}; \boldsymbol{\theta})$ that represents the implicit function (Fig.\ref{fig:Architecture}). This optimization involves minimizing a loss function consisting of two components: (a) a data fitting term $L_{\text{data}}(\boldsymbol{\theta})$, which encourages the zero iso-surface of the implicit function to closely match the input point cloud, and (b) a regularizer term $L_{\text{*}}(\boldsymbol{\theta})$ that encourages surface developability in the output. In the following sections, we elaborate on these two terms, describe the network architecture, and outline our optimization procedure.
\vspace{-2mm}

\vspace{-2mm}
\paragraph{Data term.} 
\noindent Single surface reconstruction methods involve fitting neural networks \cite{Carr:ReconRadial:2001} to a single input point cloud using data losses. The samples $\mathbf{p}_j$ used for fitting are obtained by directly selecting input points from the point cloud that have reference implicit values $s_j=0$ ("on-surface point samples"), and also by perturbing points along the normals $\mathbf{p}_j = \mathbf{p_i} + \epsilon \mathbf{n_i}$ resulting in $s_j = \epsilon$ ("off-surface point samples"). In recent data-driven approaches, surface reconstruction is performed by estimating the parameters of the implicit function from a large dataset of different point clouds. DeepSDF \cite{Park:deepsdf:2019} is one such example, utilizing an auto-decoder architecture for this purpose. In our method, we follow a similar architecture.

Our surface reconstruction network takes a point $\mathbf{p}$ as input and maps it to an implicit function $f(\mathbf{p}; \boldsymbol{\theta})$. During training, the model parameters are optimized through backpropagation using point samples $\{\bp_j,s_j\}_{j=1}^K$ taken around the input point cloud \cite{Park:deepsdf:2019}. The training process initially optimizes the parameters based on the data fitting term. After achieving the fitting, the developability regularizer is introduced for fine-tuning, resulting in a surface close to being developable. The process of fitting an implicit function to an input point cloud involves penalizing the differences between estimated and reference implicit values at various sample points surrounding the input point cloud. Specifically, given $K$ point samples $\{\bp_j\}_{j=1}^K$  with associated scalar signed distance values $\{s_j\}_{j=1}^K$, the data term can be formulate as the $L_{1}$ loss or clamped $L_{1}$ loss \cite{Park:deepsdf:2019}, which aims to make the parameter estimation more sensitive to details near the surface as follows:
\vspace{-2mm}
\begin{equation}
   L_{\text{data}}(\boldsymbol{\theta}) = \sum_{j=1}^K | f(\mathbf{p}_j;\boldsymbol{\theta}) - \text{cl}(s_j, \delta)|,
\end{equation}
, where $\text{cl}(\cdot, \delta) = \text{min}(\delta, \text{max}(-\delta, \cdot))$ and $\delta$ is a clamping parameter. We set $\delta=0.01$ in our experiments.

\paragraph{Network architecture.} \label{para:arch} Our architecture is comprised of $8$ fully connected layers, each with 512 nodes, applying group normalization. To enable the computation of Hessians for second-order differentiable regularization, we explored several second-order activation functions [siLU\cite{Elfwing:SigmoidWeightedLU:2017}, geLU\cite{Elfwing:SigmoidWeightedLU:2017}, tanH, sine \cite{sitzmann:siren:2019}, and elu]. 

\paragraph{Developability regularizer term.} \label{para:reg_term} Our regularizer term is motivated by rank minimization applied to the matrix $\hat \bH$ storing the gradients and Hessian of the implicit function in Eq.\ref{eq:Imp_gauss_curv}. We experimented with all rank minimization formulations except $L_0$ as discussed in \ref{para:rank_min}:
\begin{align}
&L_{\text{H}_{NN}}(\btheta)=\sum_{i=1}^N \, \lVert  \sigma(\bH_f(\bp_i) )  \rVert_1\\
&L_{{\hat{\text{H}}_{det}}}(\btheta)=\sum_{i=1}^N 
\det( 
\hat{\bH}_f(\bp_i)) , \forall  \mbox{rank}(\hat{\bH}_f(\bp_i)) = 3
\\
&L_{\text{H}_{logdet}}(\btheta)=\sum_{i=1}^N 
\log \det( 
\bH_f^\top(\bp_i) \cdot \bH_f(\bp_i)
+  \bI ) \\
&L_{\text{H}_{PNN}}(\btheta)= \sum_{i=1}^N
\sum_{o=r+1}^{o=3} \sigma_o( 
\bH_f(\bp_i))
\end{align}
In our experiments, we set $r=1$ (i.e., we ignore the two largest singular values of the Hessian). In our experiments, we minimized the determinants of both $\hat{\text{H}}$ and $\text{H}$, and it was observed that $\hat{\text{H}}$ exhibited better developability.

\begin{figure}[t!]
  \centering
  \includegraphics[width=1.0\linewidth]{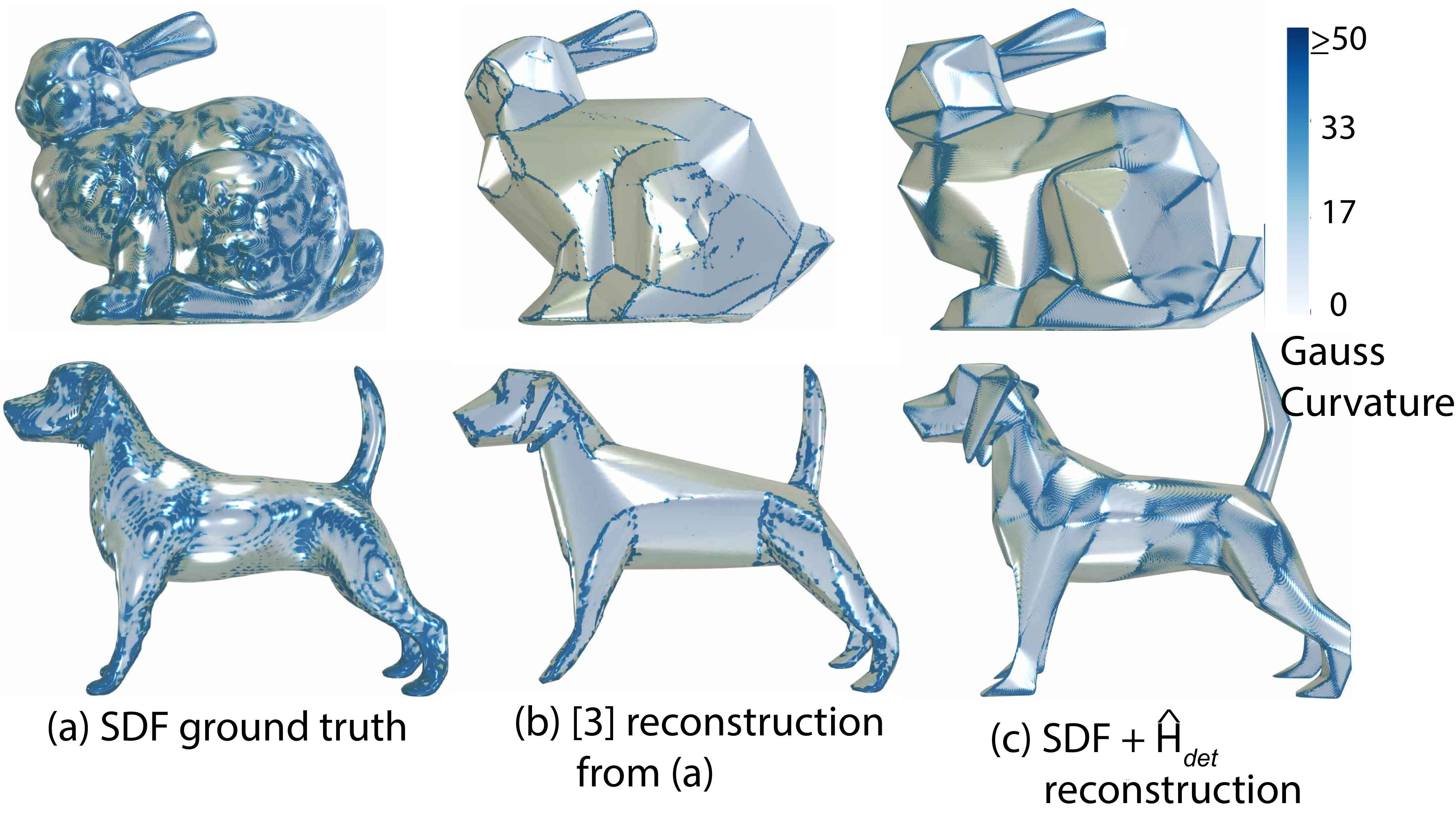}
  \vspace{-4mm}
  \caption{\label{fig:Discrete_gauss_curvature} \textbf{Visualization of Discrete Gaussian curvature of results of Table \ref{table:discrete_curvature_comparison}}. (a) SDF Ground truth(230K$\vert$57K Vertices), (b) Binninger,Verhoeven et.al \cite{BinningerVerhoeven:GaussThinning:2021} on (a), (c) marching cube \cite{Lorensen:marchingcubes:1987} reconstructed surface using $\hat{\text{H}}_{det}$ regularizer described in \ref{para:reg_term}. \emph{Note: Marching cube reconstruction lacks smooth edges and quality relies on voxel resolution (512 used here).} 
  \vspace{-2mm}
  }
\end{figure}
  
\begin{table}[t!]
\centering 
\begin{tabular}{l c c c}
\toprule
\multirow{2}{*} {}{Method} &
\multirow{1}{*} Median K	$\downarrow$ & Mean K	$\downarrow$ & CD 	$\downarrow$ \\
\midrule
\multirow{6}{*}{}
{GT} $\left|57K\right|$ & 0.004 & 0.012 & 0.0 \\
\midrule
\cite{BinningerVerhoeven:GaussThinning:2021} $\left|57K\right|$ &  $3e^{-5}$ & 0.006 & \textit{\textbf{115.1}} \\
{\cite{BinningerVerhoeven:GaussThinning:2021}} $\left|230K\right|$ &  $1e^{-5}$ & 0.003 & {\textit{\textbf{102.1}}} \\
$\text{SDF+H}_{det}\left|57K\right|$ & $1e^{-3}$ & 0.02 & 25.5\\
$\text{SDF+H}_{det}\left|950K\right|$& $5e^{-5}$ & 0.003 & 25.6\\
$\text{SDF+}\hat{\text{H}}_{det}\left|3.8M\right|$& $\boldsymbol{{1e^{-5}}}$ & \textbf{0.001} & 27.1\\

\bottomrule
\end{tabular}
\caption{\label{table:discrete_curvature_comparison}\textbf{Comparison of Discrete Gaussian curvature (K) and Chamfer Distance (CD) for bunny \ref{fig:Discrete_gauss_curvature}}. Ground-truth is the SDF surface reconstructed using marching cubes. Chamfer distance evaluated as sum of square difference of the vertices.}
\vspace{-5mm}
\end{table}

\begin{figure*}[t!]
  \centering
  \includegraphics[width=1.0\linewidth]{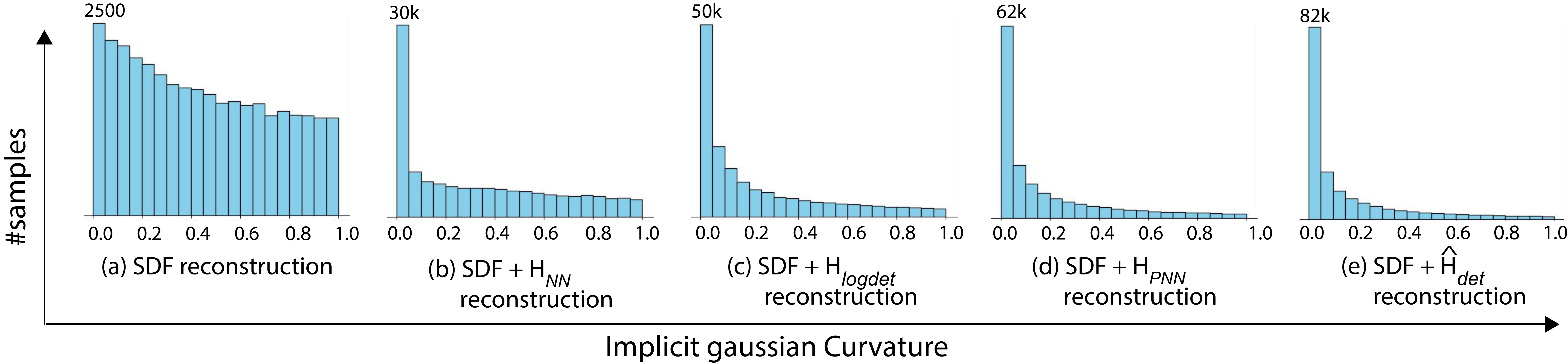}
  \vspace{-6mm}
  \caption{\label{fig:Histogram_Gauss_curvature} \textbf{Histogram of Implicit Gaussian curvature of stanford bunny} \ref{fig:reconstruction_comparison_othermodel} reconstructed using (a) SDF shape approximation without any regularizer, in comparison to the SDF Developable shape approximation using regularizer variants \ref{para:reg_term}, from (b) - (e).}
\end{figure*}

\paragraph{Minimization procedure.} Our optimization procedure aims to minimize an objective function combining the data term and developability regularizer term:
\begin{equation}
L(\btheta) = L_{\text{data}}(\btheta) + \lambda \cdot L_{*}(\btheta)
\label{eq:full_loss}
\end{equation}
 where $L_*$ denotes one of the regularizer terms detailed above(\ref{para:reg_term}), and $\lambda$ is set through hyperparameter tuning. Steps to minimizing the objective involves (1) achieving good shape approximation to the point cloud, (2) regularizing it for developability, and (3) computing its derivative with respect to the input point and network parameters. We solve the first problem by  minimizing the data term \mbox{$L_{data}(\btheta)$} for the input point cloud, yielding an approximate iso-surface. We then minimize the full loss function $L(\btheta)$ evaluated on the model parameters. Computing the implicit's Hessians needed in the developability term is feasible due to the twice-differentiable feed-forward activation function used in our network. 

We use subgradient optimization for $L_1$ norm minimization. As for other minimization terms, we leveraged pytorch's autograd (\emph{torch.autograd.grad}) to calculate the gradients and their derivatives required for backpropagation. By minimizing the complete loss function with backpropagation, we are able to obtain piecewise developable surfaces with reduced Gaussian curvature and automatic crease formation. The iterations were run until convergence.

\begin{figure}[t!]
  \centering
  \includegraphics[width=0.95\linewidth]{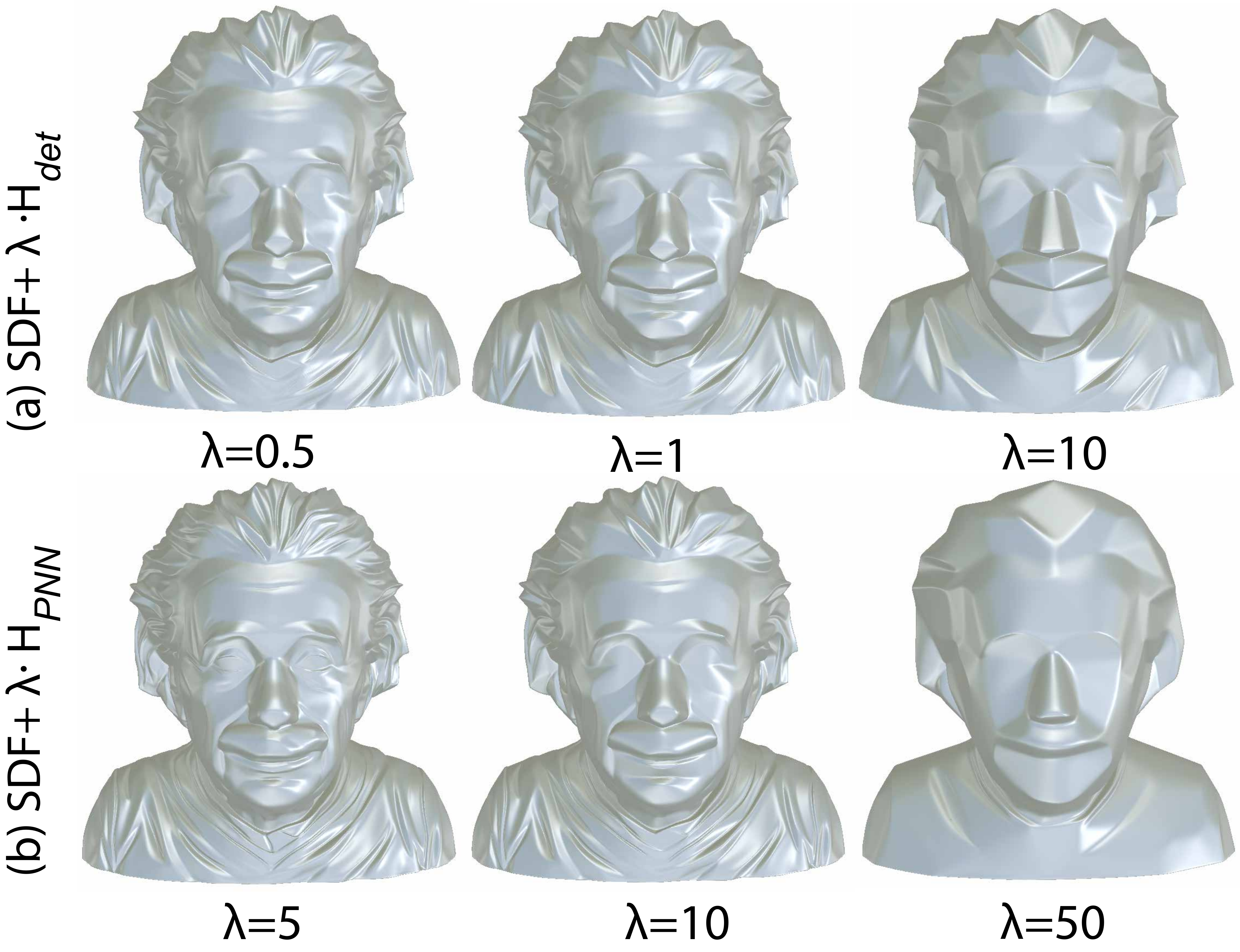}
  \vspace{-3mm}
  
  \caption{\label{fig:curvature_weight_4} \textbf{Ablation on regularizer weights, $\lambda$, and developability of surface}. Row (a), shows comparison for $\hat{\textbf{H}}_{det}$ regularized surface, row (b), shows results for $\textbf{H}_{\small{PNN}}$ regularized surface.
  \vspace{-5mm}
  }
\end{figure}

\paragraph{Implementation details.}
Prior to conducting the experiments, we standardized the inputs by fitting point cloud to unit bounding box. For shape optimization, we set the learning rate to $10^{-4}$ and adjust it to $10^{-5}$ during the fine-tuning stage for developability. We employed the Adam minimizer \cite{Kingma:Adam:2015} for optimization purposes. Our implementation is in Pytorch. The source code and data will be made available upon acceptance of the paper.

\begin{table*}
\centering
\begin{tabular}{l|c c c|c c c|c c c|c c c}
\toprule
\multirow{3}{*}{Method} & \multicolumn{3}{c|}{Bunny} & %
    \multicolumn{3}{c|}{Horse} & \multicolumn{3}{c|}{Dragon} & \multicolumn{3}{c}{Griffin}\\
\cline{2-13}
\multirow{2}{*}{}
 & \small{Med} $\downarrow$ & \small{Med} $\downarrow$ & \small{CD} $\downarrow$ & \small{Med} $\downarrow$ & \small{Med} $\downarrow$ & \small{CD} $\downarrow$ & \small{Med} $\downarrow$ & \small{Med} $\downarrow$ & \small{CD} $\downarrow$ &
 \small{Med} $\downarrow$ & \small{Med} $\downarrow$ & \small{CD} $\downarrow$\\
 & $\text{K}_{min}$ & K &  & $\text{K}_{min}$ & K &  & $\text{K}_{min}$ & K & & $\text{K}_{min}$ & K &  \\
\cline{1-13}
SDF& 1.6 & 10.2 & 14.4 & 2.0 & 19.9 & 1.3 & 2.9 & 37.4 & 2.9 & 1.6 & 18.5 & 2.5\\
SDF\cite{BinningerVerhoeven:GaussThinning:2021}& 0.2 & 0.8 & \textit{\textbf{430.7}} & \textbf{0.2} & \textbf{1.27} & \textit{\textbf{170.5}} & \textbf{0.3} & 1.8 & \textit{\textbf{860.2}} & \textbf{0.2}& 0.9 & \textit{\textbf{174.4}}\\
SDF\cite{Odedstein:DevelopabilityTriangleMeshes:2018}($\left|3K\right|$)& 0.5 & 1.3 & 54.2 & 2.0 & 18.2 & 1.6 & - & - & - & - & - &-\\
SDF+$\text{H}_{NN}$ & 0.3 & 0.8 & 71.6 & 1.4 & 9.4 & 5.0 & 1.7 & 12.7 & 75.8 & 0.8 & 5.2 & 19.4\\
SDF+$\text{H}_{logdet}$ & 0.7 & 1.7 & 62.9 & 1.2 & 8.8 & 3.9 & 1.4 & 12.1 & 5.4 & 0.6 & 3.7 & 4.0\\
SDF+$\hat{\text{H}}_{det}$ & 0.08& 0.5& 110.4 & \textbf{0.2} & \textbf{1.31} & \textbf{18.6} & 0.5 & 3.5 & 51.0 & \textbf{0.2} & \textbf{0.8} & \textbf{11.5}\\
SDF+$\text{H}_{PNN}$ & \textbf{0.03} & \textbf{0.2} & \textbf{134.7} & 0.3 & 2.1 & 10.2 & \textbf{0.4} & 4.5 & \textbf{63.2} & \textbf{0.2} & 1.7 & 15.6\\
\bottomrule
\end{tabular}

\caption{\label{table:Table_implicit_curvature_comparison} 
\textbf{Comparison of Implicit Curvature and Chamfer distance}. Implicit Curvature(K) is calculated according to Eq.\ref{eq:Imp_gauss_curv}, and $\text{K}_{min}$ is calculated using Eq.\ref{eq:principal_curvatures}, and their median value is taken (Med). Chamfer distance (CD) uses sum of the squared distance metric for 500K points. $\downarrow$ means lower values are better. For comparison, we employ SDF reconstruction from the developable surface results of other methods (SDF\cite{BinningerVerhoeven:GaussThinning:2021}, SDF\cite{Odedstein:DevelopabilityTriangleMeshes:2018}). \emph{Note: Only the implicit curvature values are measured from the SDF reconstruction, Chamfer distance is measured from their discrete representation by trying to match or lower the} Med($\text{K}_{min}$)}.
\vspace{-3mm}
\end{table*}

\begin{table}[t!]
\vspace{-2mm}
\centering 
\begin{tabular}{c c c c c}
\toprule
\multirow{2}{*} {}{Model} &
\multirow{2}{*} {}{Reg} & Med 	$\downarrow$ & Mean 	$\downarrow$ & CD 	$\downarrow$ \\
$\lambda=0$& $\lambda$ & K & K & \\
\midrule
\multirow{6}{*}{}
\multirow{6}{*}{\includegraphics[width=0.14\textwidth]{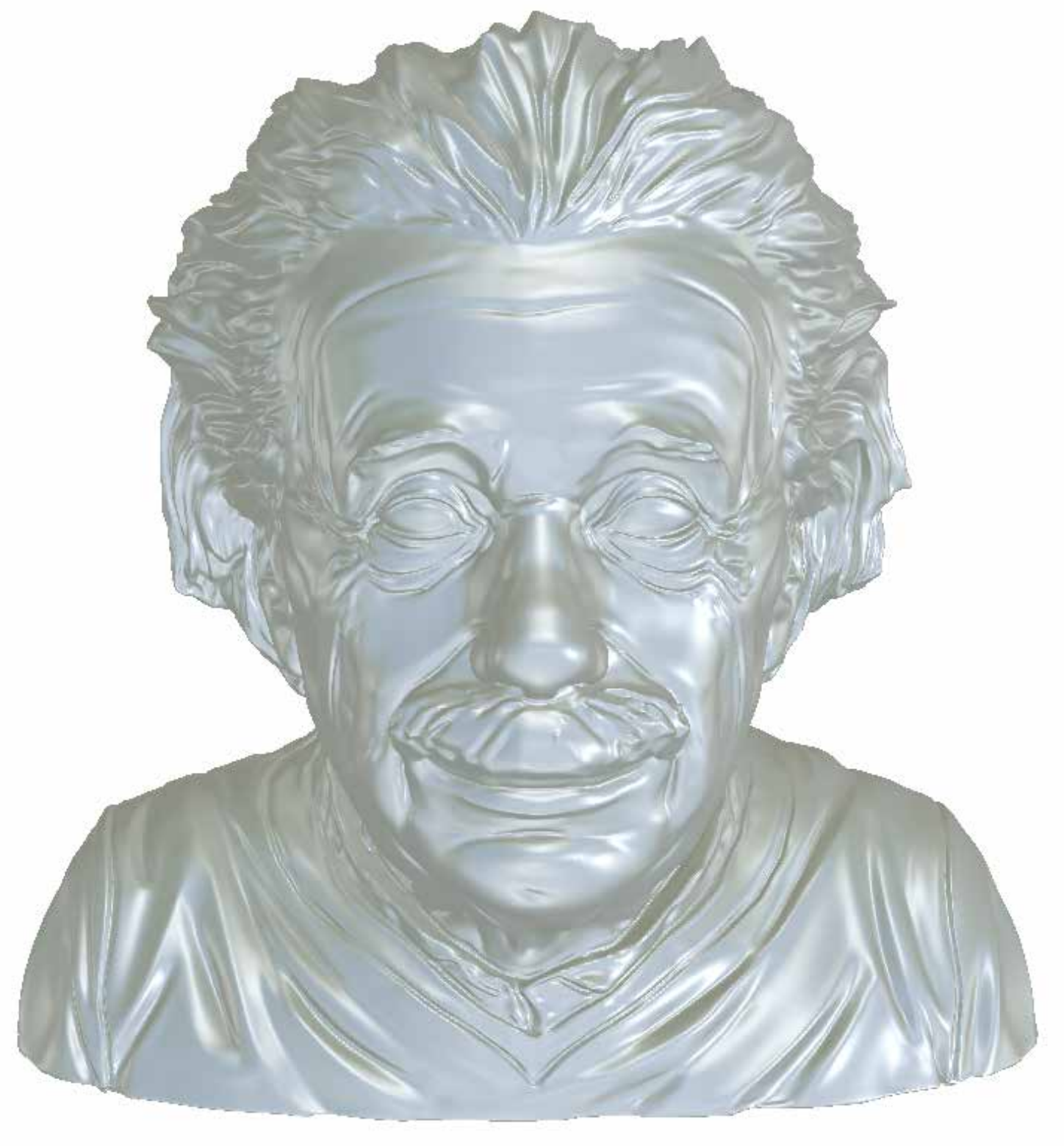}}&0 &  14.5 & 2879.2 & \textbf{18.9} \\
&0.1&   8.1 & 1227.9 & 20.7 \\
&0.5& 3.9 & 1025.3 & 25.8\\
&1&  2.6 & 625.1 & 29.7 \\
&10&  0.7 & 158.9 & 64.4 \\
&100& \textbf{0.3} & \textbf{68.2} &  166.9 \\
\bottomrule
\end{tabular}
\vspace{1mm}

\caption{\label{table:regularizer_weight_comparison} 
\textbf{Quantitative result: Ablation study on regularizer weight, $\lambda's$ effect on surface developability and chamfer distance}. 
Evaluation of implicit Gaussian curvature (K) for different $\lambda$ weights on Albert Einstein by iczfirzis licensed under CC BY-SA.(Fig.\ref{fig:curvature_weight_4}). Median(Med) and mean of K  as well as Chamfer Distance (CD) to ground-truth, are measured using 250k surface points after ICP \cite{Besl:ICP:1992}. The best performance is highlighted in bold.
\vspace{-5mm}
}
\end{table}

\begin{figure*}[t!]
  \centering
  \includegraphics[width=0.95\linewidth]{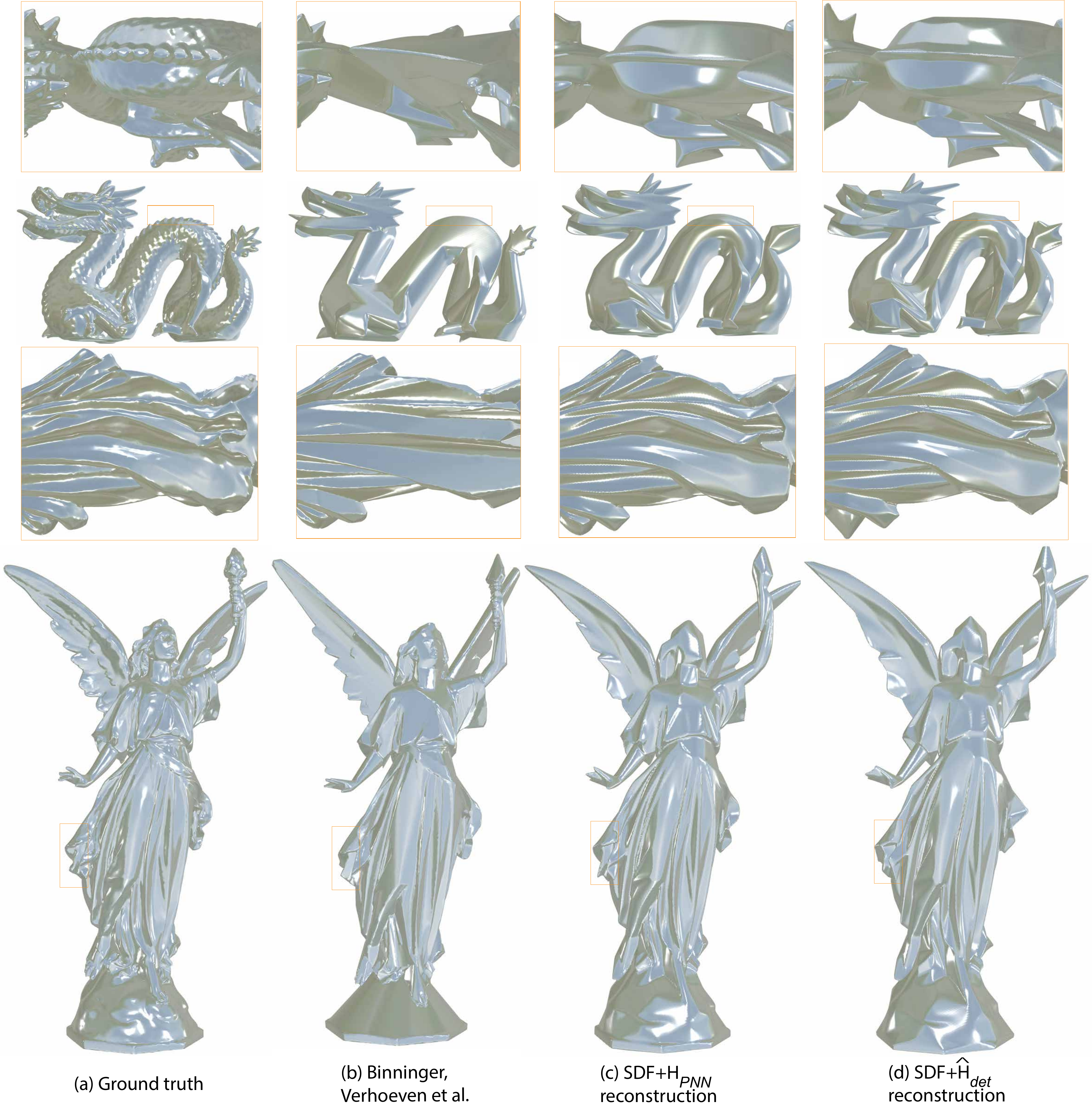}
  \caption{\label{fig:reconstruction_comparison_othermodel} \textbf{Comparison of developable surface reconstruction} results of two of our regularization methods (c) and (d) from \ref{para:reg_term} with the results of (b) Binninger,Verhoeven et al. \cite{BinningerVerhoeven:GaussThinning:2021}. The insets shows that (b) fails to preserve the structure of the shape while our methods does. Row 1 shows the rotated image of back of the dragon, Row 3 shows drapes of back side of Lucy, Row 4 the wings}
\end{figure*}

%% file: sec/5_experiments.tex
\section{Experiments}

We now discuss our experiments for evaluating the effects of the proposed developability regularization, the evaluation metrics, and finally show results and comparisons.

\vspace{-2mm}
\paragraph{Competing variants.}  We evaluate different variants based on the same network architecture mentioned in Section \ref{para:arch}. The evaluated variants include:

(a) Activation variants: \label{para:activation} \emph{geLU}, \emph{siLU}, \emph{tanh}, \emph{elu} used in the network for both shape approximation and fine-tuning for developability (ref. \emph{supplement})

(b) Regularization variants: $\text{H}_{\small{NN}}$, $\text{H}_{logdet}$, $\text{H}_{PNN}$, $\hat{\text{H}}_{det}$, which perform fine-tuning of the network parameters based on the total loss function (Eq.\ref{eq:full_loss}). $\text{H}_{NN}$, $\text{H}_{logdet}$ and $\text{H}_{PNN}$  minimizes the $L_1$ norm of the singular values, the logarithmic determinant of squared hessian and the lowest singular value of the implicit hessian $\bH_f(\bp)$ respectively, while $\hat{\text{H}}_{det}$  minimizes the determinant of the matrix $\hat{\bH}_f(\bp)$ (Eq. \ref{eq:dethath}). For fair comparison, these variants were evaluated with the geLU activation function.

\vspace{-2mm}
\paragraph{Evaluation metrics.} 
We evaluate the regularizer variants using two metrics. Firstly, we employ the Chamfer distance metric to assess the similarity between the reconstructed surface and the ground-truth surface. This involves sampling 250k points on both surfaces (using Poisson disk uniform sampling) followed by iterative closest point (ICP) \cite{Besl:ICP:1992} for alignment. Then bidirectional Chamfer distance is calculated between the point samples.
Secondly, to measure surface developability, we compute the median of the absolute minimum of the implicit principal curvatures ($K_\text{min}$) at each point. Measuring the minimum is crucial in cases where either one of $K_1$ or $K_2$ is close to 0 while the other possesses a high value (such as creases). This allows us to accurately assess the overall curvature, avoiding underestimation of developability at points where the surface is nearly flat. To compute the principal curvatures, and we follow Goldman \cite{Goldman:CurvatureImplicit:2005} to obtain the implicit mean curvature $M$:
\vspace{-1mm}
\begin{equation}
M(\bold{p})\!\!=\!\!
\frac{
\nabla {f(\bp)} \cdot \bH_f(\bp) \cdot \nabla {f(\bp)^T} - \left| {\nabla f(\bp)} \right|^2 \cdot \text{Tr}(\bH_f(\bp))}
{2{
\left| {\nabla f(\bp)} \right|^3
}}
\label{eq:Imp_mean_curv}
\end{equation}

\noindent Using Eq.\ref{eq:Imp_gauss_curv} and Eq.\ref{eq:Imp_mean_curv}, we get the principal curvatures values and their absolute minimum as follows  \cite{Goldman:CurvatureImplicit:2005} :
\begin{align}
    &K_1, K_2 \!\!=\!\!
    M(\bp) \pm \sqrt{M^2(\bp) - K(\bp)}\\
    &K_{\text{min}}\!\!=\!\!
    \text{min}(\left|K_1\right|,\left|K_2\right|)
    \label{eq:principal_curvatures}
\end{align}

\paragraph{Results.}{\label{para:results}}
Tables \ref{table:discrete_curvature_comparison}, \ref{table:Table_implicit_curvature_comparison} and \ref{table:regularizer_weight_comparison} present our quantitative evaluation of the regularizer variants. We find that all variants incorporating our developability term significantly reduce Gaussian curvature on the reconstructed surface compared to the ground truth. It is worth noting that while approaching developability, the shape approximation of the reconstruction does not deviate much(Fig.\ref{fig:reconstruction_comparison_othermodel}). It shown in the Chamfer distance metric compared to other methods (\ref{table:discrete_curvature_comparison}, \ref{table:Table_implicit_curvature_comparison}). Additionally, we observe that each variant offers a different level of developability approximation (Fig.\ref{fig:teaser}), owing to the varied approaches and relaxations used to minimize the hessian rank, but the $\text{H}_{NN}$ and $\hat{\text{H}}_{det}$ variants provide consistent piecewise developable patches with lower $K_\text{min}$ and lower chamfer distance. The same conclusion can be observed qualitatively (\ref{fig:teaser},\ref{fig:Discrete_gauss_curvature}). $\hat{\text{H}}_{det}$ gives piecewise planar developable patches, $\text{H}_{NN}$ gives planar and non-planar developable patches, $\text{H}_{logdet}$ minimizes overall curvature, but produces stronger crease lines, while $\text{H}_{NN}$ tries to minimize both the principal curvatures but has adverse effect on shape approximation.

\begin{figure}[t!]
  \centering
  \includegraphics[width=1.0\linewidth]{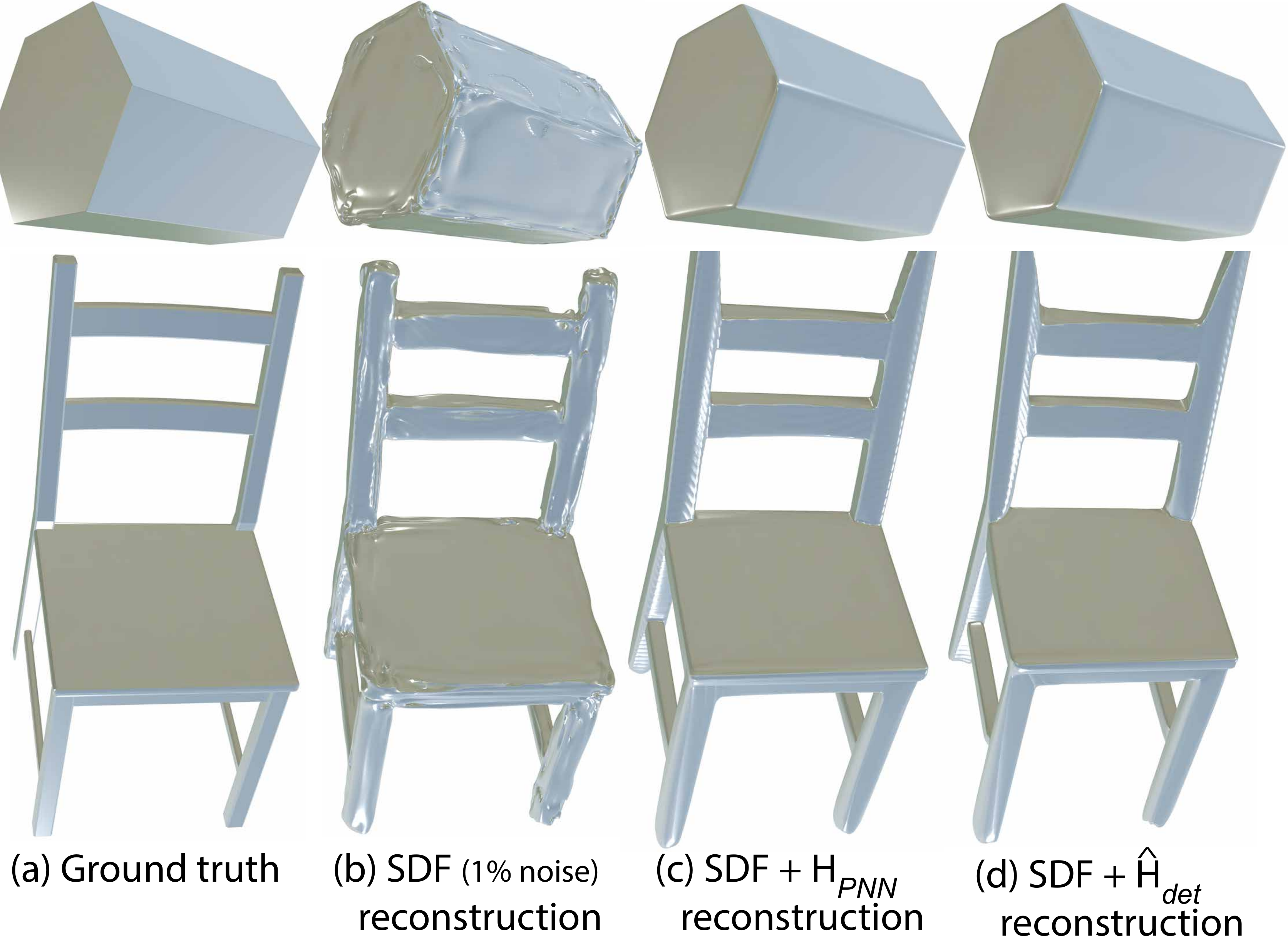}
  \vspace{-4mm}
  \caption{\label{fig:Noise_robustness} \textbf{Comparison of developbale surface reconstruction results with noisy input} point cloud ($1\%$ noise) for and $\text{H}_{PNN}$ and $\hat{\text{H}}_{det}$ regularizer variants.}
  \vspace{-2mm}
\end{figure}

\vspace{-2mm}
\paragraph{Robustness.} We introduce a perturbation of $1\%$ to the positions of the input point cloud while preserving the normal sign. In Figure \ref{fig:Noise_robustness}, (b) shows the reconstruction of noisy point cloud using SDF reconstruction without any regularizer, while (c) and (d) depict the reconstructions using the $\text{H}_{NN}$ and $\hat{\text{H}}_{det}$ regularizer variants for the same noisy input. Despite the noise, the reconstructed surfaces remain close to the ground truth surface. Note that, as the noise $\%$ increases, the surface SDF reconstruction gets thicker.

\vspace{-2mm}
\paragraph{Ablation Results.}
Our findings reveal that increasing the regularizer weight enhances developability, as evident from the lower median and mean curvatures shown in Table  \ref{table:regularizer_weight_comparison}. However, this improvement comes with the trade-off of a greater deviation from the shape approximation.

%% file: sec/7_conclusion.tex
\section{Conclusion}
We have developed a method to approximate the developability of surfaces, applicable to closed surfaces with varying levels of detail. Our approach leverages the implicit representation of the surface and introduces a novel regularizer term that acts on the implicit's hessian and gradient, encouraging the emergence of a piecewise developable surface with automatic crease formation. Experimental results demonstrate the results of developability achieved by our method while preserving the surface's structural characteristics better than alternative techniques.

\begin{figure}[t!]
  \centering
  \includegraphics[width=0.8\linewidth]{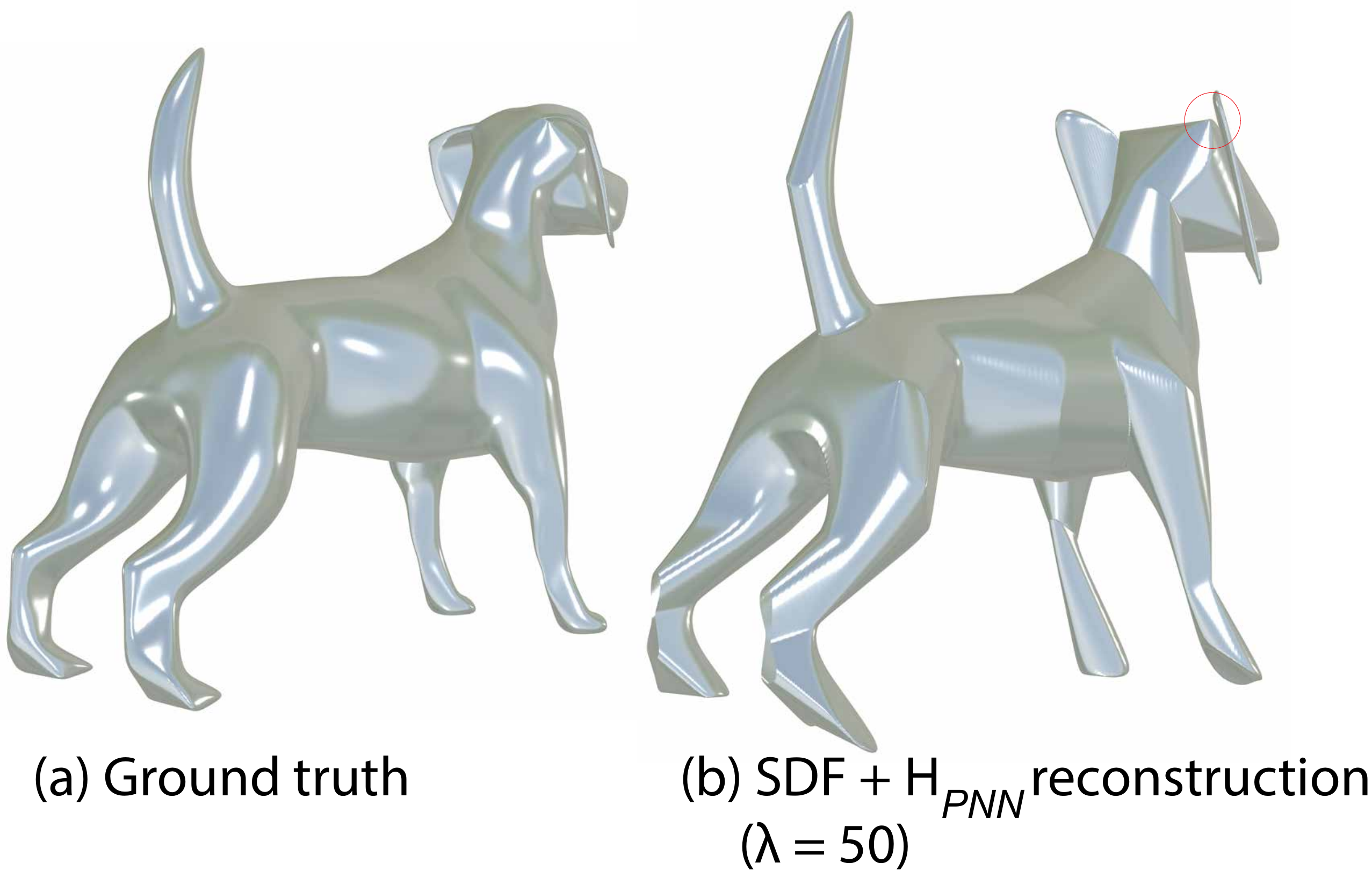}
  \vspace{-1mm}
  \caption{\label{fig:failure}\textbf{Failed case using our method.} (b) shows the reconstruction using $\text{H}_{PNN}$ regularizer variant using larger regularizer weight $\lambda$. Thin connectivity structure (dog ears) gets disconnected or vanishes.
  \vspace{-3mm}
  }
\end{figure}
\vspace{-2mm}
\paragraph{Limitations and future work.}
While our method shows promising results, it works only on single closed surfaces and assumes correct normal signs in the provided point cloud. With the current architecture the natural next step is to generalize this as data-driven approach while extending it to handle open surfaces and noisy normals. Our method is a global shape optimization method, thus does not adequately preserve details in regions in a shape with different levels of detailing. (Fig.\ref{fig:reconstruction_comparison_othermodel}, Lucy model, where higher weights makes drapes with more developable, but face details are not preserved). Moreover, there are potential failures arising from marching cube reconstruction, which may not ensure connectivity for shapes with thinner connectivity structures, as illustrated in Figure \ref{fig:failure} with higher regularizer weights. Thus, automatically segmenting point clouds and applying appropriate regularizer for individual segments is a promising avenue for future research.

%% file: sec/X_suppl.tex
\clearpage
\setcounter{page}{1}
\maketitlesupplementary

\section{Implicit Curvature comparison}
\label{sec:SDF}

In Table.\ref{table:Table_implicit_curvature_comparison}, we compared the implicit curvature metric between the state-of-the-art developable surface methods which operates on discrete representation and our method. To facilitate this evaluation, we transformed the discrete output to implicit representation by sampling 250k points from their surface and fitted it using a SDF. To validate this evaluation approach, the following table \ref{table:Table_SDF_original_comparison} showcases the calculated chamfer distances between the sampled points from the discrete surfaces and those from the SDF-reconstructed surfaces.
\begin{table}[htb]
\centering 
\begin{tabular}{l l c}
\toprule
\multirow{2}{*} {}Model &
\multirow{2}{*} {}Method & CD 	$\downarrow$ \\
 &  &($\text{n}_{samples}$ = 250k)\\
\midrule
\multirow{2}{*}
 {Bunny} & SDF\cite{BinningerVerhoeven:GaussThinning:2021}   & 3.5 \\
 & SDF\cite{Odedstein:DevelopabilityTriangleMeshes:2018} & 4.4 \\
  \midrule
 \multirow{2}{*}
 {Horse} & SDF\cite{BinningerVerhoeven:GaussThinning:2021}    & 1.3 \\
 & SDF\cite{Odedstein:DevelopabilityTriangleMeshes:2018} & 2.5 \\
 \midrule
 {Dragon} & SDF\cite{BinningerVerhoeven:GaussThinning:2021} & 3.8 \\
 \midrule
 {Griffin} & SDF\cite{BinningerVerhoeven:GaussThinning:2021} & 2.2 \\

\bottomrule
\end{tabular}
\vspace{-2mm}
\caption{\label{table:Table_SDF_original_comparison} 
\textbf{Chamfer distance(CD) comparison of Sellan et al.\cite{Sellan:Developability:2020} and Binninger,Verhoeven et al.\cite{BinningerVerhoeven:GaussThinning:2021} with its corresponding SDF reconstruction}. Grid resolution of 512 is used for for marching cubes \cite{Lorensen:marchingcubes:1987} reconstruction.
}
\end{table}

\begin{figure*}[tbh]
  
  \centering
  \includegraphics[width=1.0\linewidth]{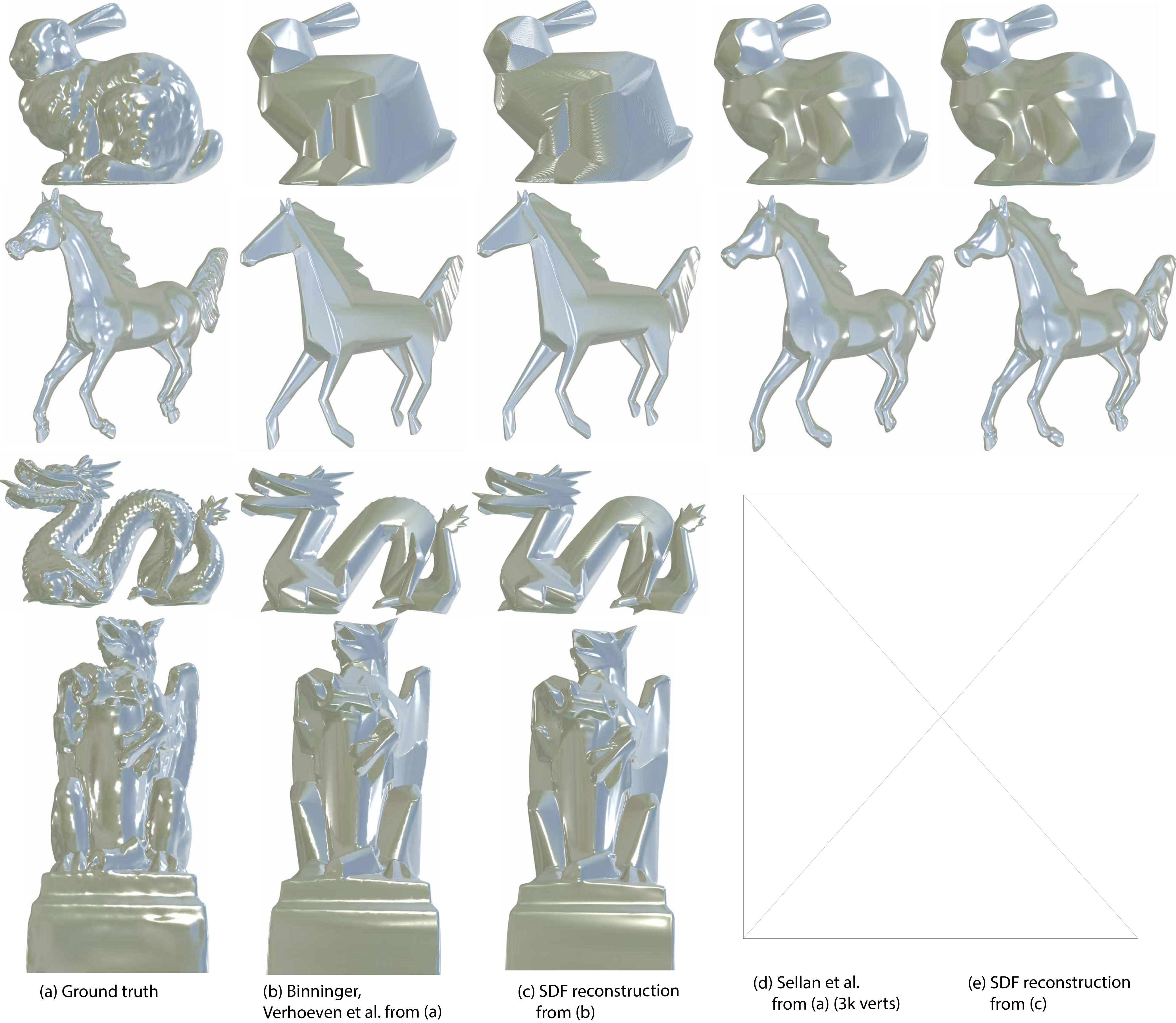}
  \vspace{-4mm}
  \caption{\label{fig:sdf_original_comparison} \textbf{Comparison of discrete vs SDF reconstructed developable surfaces of \cite{BinningerVerhoeven:GaussThinning:2021} and \cite{Sellan:Developability:2020}.} (a) shows the Ground truth, (b) shows developable surface constructed using Binninger,Verhoeven et al. \cite{BinningerVerhoeven:GaussThinning:2021} with maximum of 50k vertices, (d) shows developable surface constructed using Sellen et al. \cite{Sellan:Developability:2020} with maximum of $\approx$3k vertices, (c) and (e) shows corresponding SDF reconstructed surface from (b) and (d) as ground truth.
  \vspace{-2mm}
  }
\end{figure*}

Note that Sellan et al. \cite{Sellan:Developability:2020} required a reduced polygon count for compilation and execution. Consequently, it was only tested on Bunny and Horse models, achieved by reducing polygon numbers through meshlab \cite{Cignoni:Meshlab:2008}. Dragon and Griffin models were excluded from this method, since reduction of polygons led to loss of their surface details. The lower chamfer distance in Table \ref{table:Table_SDF_original_comparison} indicates that the comparison of implicit curvature of our method against SOTA methods in Table \ref{table:Table_implicit_curvature_comparison} demonstrates minimal deviation from the ground truth discrete representation. The quantitative outcome is also illustrated in Figure \ref{fig:sdf_original_comparison}.

 \begin{figure*}
  \centering
  \includegraphics[width=1.0\linewidth]{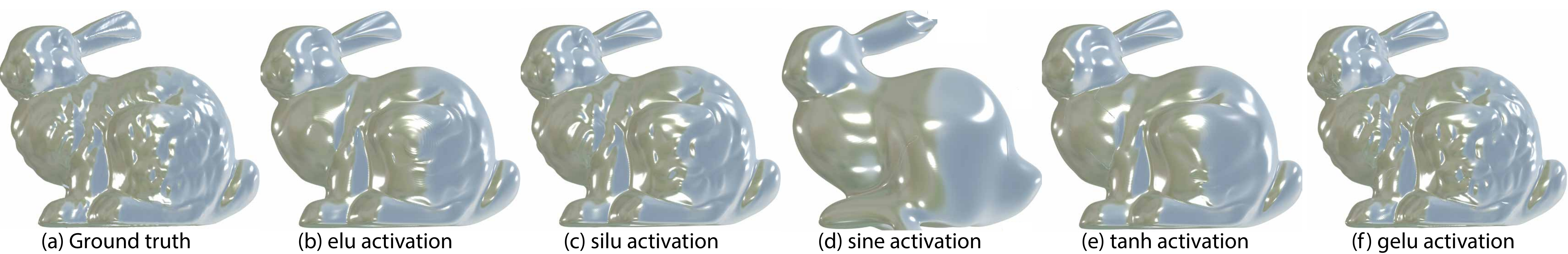}
  \vspace{-4mm}
  \caption{\label{fig:Activation_comparison} \textbf{SDF reconstruction with different activation functions}. (b) - (f) shows the SDF reconstruction of stanford bunny using second order differentiable activation functions under same constraints of epoch and learning rate(Ref. \ref{para:activation}).
  \vspace{-2mm}
  }
\end{figure*}

\section{Activation comparison}
We explored several second-order activation functions for SDF reconstruction. The objective was to achieve both improved reconstruction accuracy in comparison to the ground truth and faster convergence. During experimentation, a learning rate of $1e^{-4}$ was used, and the training process was executed until convergence. The $gelu$ activation function exhibited faster convergence and better alignment with the ground truth, as assessed through the chamfer distance metric. These results are depicted in Figure \ref{fig:Activation_comparison}.

\section{Developable surface results comparison}
\begin{figure*}[t!]
  \centering
  \includegraphics[width=1.0\linewidth]{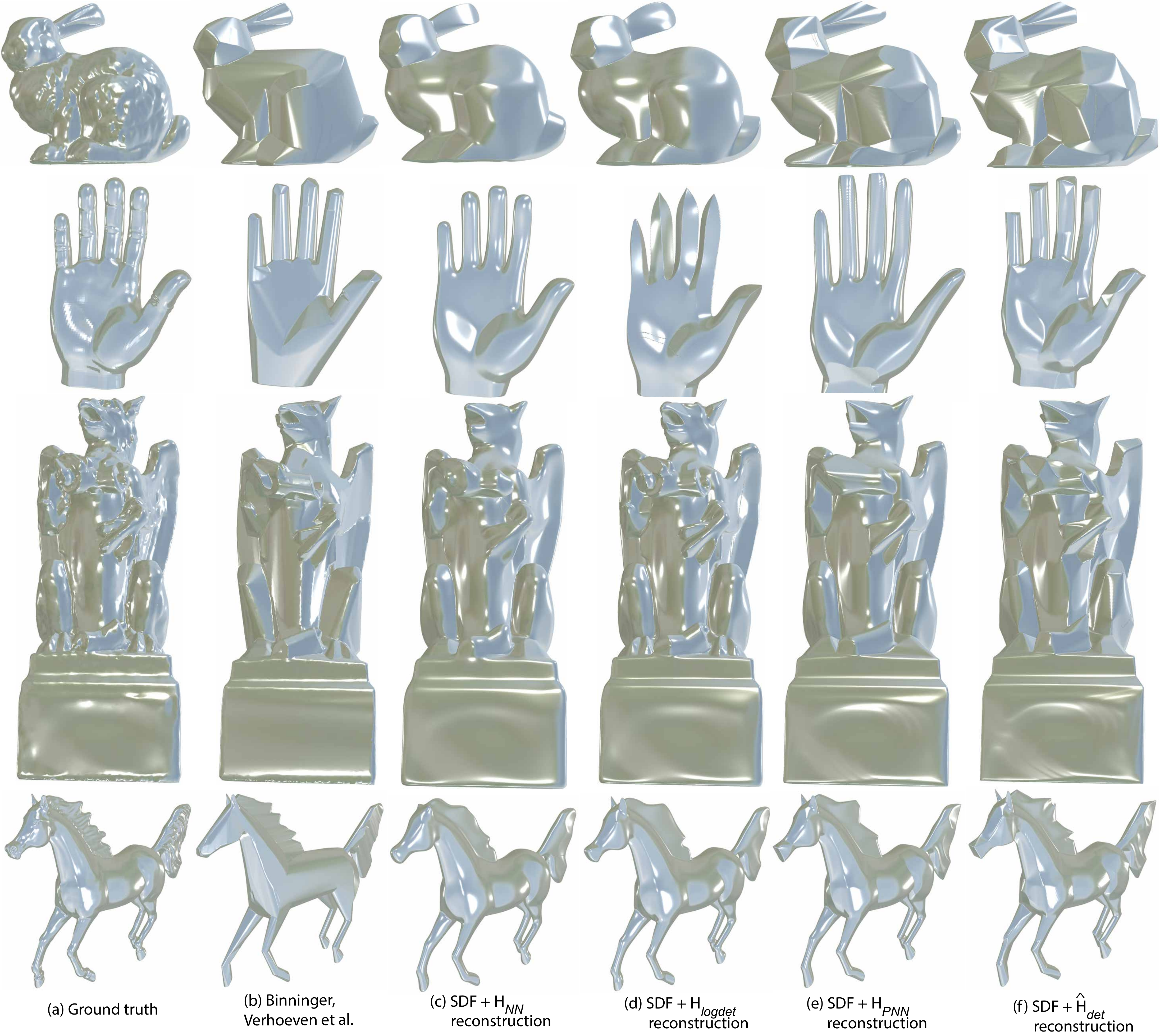}
  \vspace{-4mm}
  \caption{\label{fig:supp_comparison} \textbf{Comparison of developable surface reconstruction} results of our regularization methods (c) - (f) (Ref.\ref{para:reg_term}) with the results of (b) Binninger,Verhoeven et al. \cite{BinningerVerhoeven:GaussThinning:2021}
  \vspace{-3mm}
  }
\end{figure*}

\begin{figure*}[t!]
  \centering
  \includegraphics[width=1.0\linewidth]{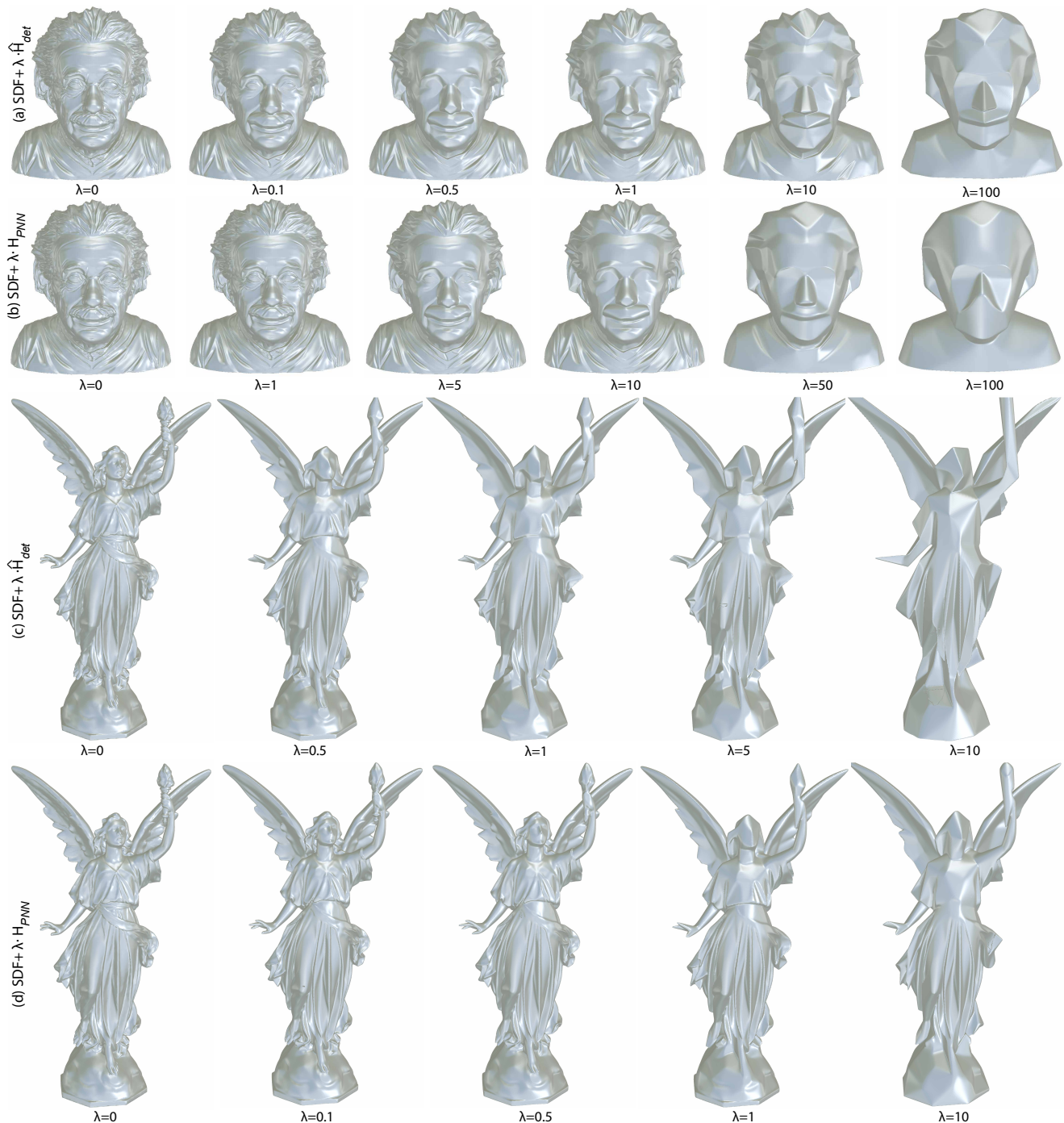}
  \vspace{-4mm}
  \caption{\label{fig:supp_curvature_weight}
  \textbf{Ablation on regularizer weights, $\lambda$, and developability of surface}. Row (a) and (c), shows comparison for $\hat{\text{H}}_{det}$ regularized surface, row (b) and (d), shows results for $\text{H}_{\small{PNN}}$ regularized surface. As the weight increases, the surfaces gain more developability, which is a desired characteristic. However, this improvement comes at the expense of preserving the original shape structure.
  \vspace{-10mm}
  }
\end{figure*}

Figure.\ref{fig:supp_comparison} shows additional comparison of developable surface constructed by Binninger,Verhoeven et al.\cite{BinningerVerhoeven:GaussThinning:2021} and the implicit reconstruction from our method using all the variants described in Section.\ref{sec:method}.
Figure.\ref{fig:supp_curvature_weight} shows the developable surface results varying the weight of the regularizer for $\text{H}_{PNN}$ and $\hat{\text{H}}_{det}$ variant from our method.

%% file: arxiv.bbl
\begin{thebibliography}{42}
\providecommand{\natexlab}[1]{#1}
\providecommand{\url}[1]{\texttt{#1}}
\expandafter\ifx\csname urlstyle\endcsname\relax
  \providecommand{\doi}[1]{doi: #1}\else
  \providecommand{\doi}{doi: \begingroup \urlstyle{rm}\Url}\fi

\bibitem[Amos et~al.(2020)Amos, Lior, Niv, Matan, and
  Yaron]{Gropp:ImplicitGeometricRegularization:2020}
Gropp Amos, Yariv Lior, Haim Niv, Atzmon Matan, and Lipman Yaron.
\newblock Implicit geometric regularization for learning shapes.
\newblock In \emph{Proceedings of the 37th International Conference on Machine
  Learning}. JMLR.org, 2020.

\bibitem[April and John~E.(2021)]{sagan:Lowrankfactor:2021}
Sagan April and Mitchell John~E.
\newblock Low-rank factorization for rank minimization with nonconvex
  regularizers.
\newblock \emph{Computational Optimization and Applications}, 79\penalty0
  (2):\penalty0 273--300, 2021.

\bibitem[Binninger et~al.(2021)Binninger, Verhoeven, Herholz, and
  Sorkine-Hornung]{BinningerVerhoeven:GaussThinning:2021}
Alexandre Binninger, Floor Verhoeven, Philipp Herholz, and Olga
  Sorkine-Hornung.
\newblock Developable approximation via gauss image thinning.
\newblock \emph{Computer Graphics Forum (proceedings of SGP 2021)}, 40\penalty0
  (5):\penalty0 289--300, 2021.

\bibitem[Carr et~al.(2001)Carr, Beatson, Cherrie, Mitchell, Fright, McCallum,
  and Evans]{Carr:ReconRadial:2001}
J.~C. Carr, R.~K. Beatson, J.~B. Cherrie, T.~J. Mitchell, W.~R. Fright, B.~C.
  McCallum, and T.~R. Evans.
\newblock Reconstruction and representation of 3d objects with radial basis
  functions.
\newblock In \emph{Proceedings of the 28th Annual Conference on Computer
  Graphics and Interactive Techniques}, page 67–76. Association for Computing
  Machinery, 2001.

\bibitem[Charlie and Kai(2004)]{Wang:Achievingdevelopability:2004}
Wang Charlie and Tang Kai.
\newblock Achieving developability of a polygonal surface by minimum
  deformation: A study of global and local optimization approaches.
\newblock \emph{Visual Computer}, 20:\penalty0 521--539, 2004.

\bibitem[Chengcheng et~al.(2016)Chengcheng, Pengbo, Johannes, and
  Helmut]{Tang:InteractiveDesign:2016}
Tang Chengcheng, Bo Pengbo, Wallner Johannes, and Pottmann Helmut.
\newblock Interactive design of developable surfaces.
\newblock \emph{ACM Trans. Graph.}, 35\penalty0 (2), 2016.

\bibitem[Chong et~al.(2015)Chong, Zhao, Huiqing, and
  Qiang]{Peng:Subspacecluster:2015}
Peng Chong, Kang Zhao, Li Huiqing, and Cheng Qiang.
\newblock Subspace clustering using log-determinant rank approximation.
\newblock In \emph{Proceedings of the 21th ACM SIGKDD International Conference
  on Knowledge Discovery and Data Mining}, page 925–934, New York, NY, USA,
  2015. Association for Computing Machinery.

\bibitem[Christos et~al.(2018)Christos, Max, and
  Diederik]{Louizos:l0learning:2018}
Louizos Christos, Welling Max, and P.~Kingma Diederik.
\newblock Learning sparse neural networks through l\_0 regularization.
\newblock In \emph{International Conference on Learning Representations}, 2018.

\bibitem[David and Bhaskar(2004)]{Wipf:L0min:2004}
Wipf David and Rao Bhaskar.
\newblock L0-norm minimization for basis selection.
\newblock In \emph{Proceedings of the 17th International Conference on Neural
  Information Processing Systems}, page 1513–1520, Cambridge, MA, USA, 2004.
  MIT Press.

\bibitem[Diederik~P. and Jimmy(2015)]{Kingma:Adam:2015}
Kingma Diederik~P. and Ba Jimmy.
\newblock Adam: {A} method for stochastic optimization.
\newblock In \emph{3rd International Conference on Learning Representations,
  {ICLR} 2015, San Diego, CA, USA, May 7-9, 2015, Conference Track
  Proceedings}, 2015.

\bibitem[Hsueh-Ti~Derek and Alec(2019)]{Liu:CubicStyle:2019}
Liu Hsueh-Ti~Derek and Jacobson Alec.
\newblock Cubic stylization.
\newblock \emph{ACM Trans. Graph.}, 38\penalty0 (6), 2019.

\bibitem[Idan et~al.(2006)Idan, Ayellet, and
  George]{Shatz:PaperCraftModels:2006}
Shatz Idan, Tal Ayellet, and Leifman George.
\newblock Paper craft models from meshes.
\newblock \emph{Vis. Comput.}, 22\penalty0 (9):\penalty0 825–834, 2006.

\bibitem[Ion et~al.(2020)Ion, Rabinovich, Herholz, and
  Sorkine{-}Hornung]{Ion:ApproximatingDOGs:2020}
Alexandra Ion, Michael Rabinovich, Philipp Herholz, and Olga Sorkine{-}Hornung.
\newblock Shape approximation by developable wrapping.
\newblock \emph{ACM Transactions on Graphics (proceedings of SIGGRAPH ASIA)},
  39\penalty0 (6), 2020.

\bibitem[Jeong~Joon et~al.(2019)Jeong~Joon, Peter, Julian, Richard, and
  Steven]{Park:deepsdf:2019}
Park Jeong~Joon, Florence Peter, Straub Julian, Newcombe Richard, and Lovegrove
  Steven.
\newblock Deepsdf: Learning continuous signed distance functions for shape
  representation.
\newblock In \emph{The IEEE Conference on Computer Vision and Pattern
  Recognition (CVPR)}, 2019.

\bibitem[Jun and Hiromasa(2004)]{Mitani:MakingPapercraft:2004}
Mitani Jun and Suzuki Hiromasa.
\newblock Making papercraft toys from meshes using strip-based approximate
  unfolding.
\newblock \emph{ACM Trans. Graph.}, 23\penalty0 (3):\penalty0 259–263, 2004.

\bibitem[Justin et~al.(2012)Justin, Etienne, Max, and
  Eitan]{Solomon:FlexibleDevSurf:2012}
Solomon Justin, Vouga Etienne, Wardetzky Max, and Grinspun Eitan.
\newblock {Flexible Developable Surfaces}.
\newblock \emph{Computer Graphics Forum}, 2012.

\bibitem[Keenan et~al.(2013)Keenan, Fernando, Mathieu, and
  Peter]{Crane:DGP:2013}
Crane Keenan, de~Goes Fernando, Desbrun Mathieu, and Schröder Peter.
\newblock Digital geometry processing with discrete exterior calculus.
\newblock In \emph{ACM SIGGRAPH 2013 Courses}. Association for Computing
  Machinery, 2013.

\bibitem[Kenneth et~al.(2007)Kenneth, Alla, Jamie, Marie-Paule, and
  Boris]{Kenneth:DevSurfSketchBound:2007}
Rose Kenneth, Sheffer Alla, Wither Jamie, Cani Marie-Paule, and Thibert Boris.
\newblock Developable surfaces from arbitrary sketched boundaries.
\newblock In \emph{Proceedings of the Fifth Eurographics Symposium on Geometry
  Processing}, page 163–172. Eurographics Association, 2007.

\bibitem[Konstantinos et~al.(2019)Konstantinos, Alexander, and
  Helmut]{Gavriil:OptimizingBSpline:2019}
Gavriil Konstantinos, Schiftner Alexander, and Pottmann Helmut.
\newblock Optimizing b-spline surfaces for developability and paneling
  architectural freeform surfaces.
\newblock \emph{Computer-Aided Design}, 111:\penalty0 29--43, 2019.

\bibitem[Li et~al.(2011)Li, Cewu, Yi, and Jiaya]{Xu:Imagesmoothl0:2011}
Xu Li, Lu Cewu, Xu Yi, and Jia Jiaya.
\newblock Image smoothing via l0 gradient minimization.
\newblock \emph{ACM Trans. Graph.}, 30\penalty0 (6):\penalty0 1–12, 2011.

\bibitem[Manfredo~P.(1976)]{docarmo:DiffGeom:1976}
do~Carmo Manfredo~P.
\newblock \emph{Differential Geometry of Curves and Surfaces}.
\newblock Prentice Hall, 1976.

\bibitem[Martin(2004)]{Peternell:DevSurfToPointCloud:2004}
Peternell Martin.
\newblock Developable surface fitting to point clouds.
\newblock \emph{Computer Aided Geometric Design}, 21\penalty0 (8):\penalty0
  785–803, 2004.

\bibitem[Maryam et~al.(2001)Maryam, Haitham~A., and Stephen~P.]{Fazel:ARM:2001}
Fazel Maryam, Hindi Haitham~A., and Boyd Stephen~P.
\newblock A rank minimization heuristic with application to minimum order
  system approximation.
\newblock In \emph{Proceedings of the 2001 American Control Conference. (Cat.
  No.01CH37148)}, pages 4734--4739 vol.6, 2001.

\bibitem[Maryam et~al.(2003)Maryam, Haitham~A., and
  Stephen~P.]{Fazel:LogdetHeuristic:2003}
Fazel Maryam, Hindi Haitham~A., and Boyd Stephen~P.
\newblock Log-det heuristic for matrix rank minimization with applications to
  hankel and euclidean distance matrices.
\newblock In \emph{Proceedings of the 2003 American Control Conference, 2003.},
  pages 2156--2162 vol.3, 2003.

\bibitem[Maryan(2002)]{Fazel:Thesis:2002}
Fazel Maryan.
\newblock \emph{Matrix Rank Minimization with Applications}.
\newblock PhD thesis, Department of Electrical Engineering, Stanford
  University, 2002.

\bibitem[Natarajan(1995)]{Natarajan:SparseApprox:1995}
B.~K. Natarajan.
\newblock Sparse approximate solutions to linear systems.
\newblock \emph{SIAM Journal on Computing}, 24\penalty0 (2):\penalty0 227--234,
  1995.

\bibitem[Oded et~al.(2018)Oded, Eitan, Max, and
  Alec]{Stein:NaturalBoundaryCond:2018}
Stein Oded, Grinspun Eitan, Wardetzky Max, and Jacobson Alec.
\newblock Natural boundary conditions for smoothing in geometry processing.
\newblock \emph{ACM Trans. Graph.}, 37\penalty0 (2), 2018.

\bibitem[Paolo et~al.(2008)Paolo, Marco, Massimiliano, Matteo, Fabio, and
  Guido]{Cignoni:Meshlab:2008}
Cignoni Paolo, Callieri Marco, Corsini Massimiliano, Dellepiane Matteo,
  Ganovelli Fabio, and Ranzuglia Guido.
\newblock {MeshLab: an Open-Source Mesh Processing Tool}.
\newblock In \emph{Eurographics Italian Chapter Conference}. The Eurographics
  Association, 2008.

\bibitem[P.J. and Neil~D.(1992)]{Besl:ICP:1992}
Besl P.J. and McKay Neil~D.
\newblock A method for registration of 3-d shapes.
\newblock \emph{IEEE Transactions on Pattern Analysis and Machine
  Intelligence}, 14\penalty0 (2):\penalty0 239--256, 1992.

\bibitem[Rabinovich et~al.(2018)Rabinovich, Hoffmann, and
  Sorkine-Hornung]{Rabinovich:DiscGeoNets:2018}
Michael Rabinovich, Tim Hoffmann, and Olga Sorkine-Hornung.
\newblock Discrete geodesic nets for modeling developable surfaces.
\newblock \emph{ACM Transactions on Graphics}, 37\penalty0 (2), 2018.

\bibitem[Ron(2005)]{Goldman:CurvatureImplicit:2005}
Goldman Ron.
\newblock Curvature formulas for implicit curves and surfaces.
\newblock 22:\penalty0 632--658, 2005.

\bibitem[Silvia et~al.(2020)Silvia, Noam, and Alec]{Sellan:Developability:2020}
Sellán Silvia, Aigerman Noam, and Jacobson Alec.
\newblock Developability of heightfields via rank minimization.
\newblock \emph{ACM Transactions on Graphics}, 2020.

\bibitem[Stefan et~al.(2017)Stefan, Eiji, and
  Kenji]{Elfwing:SigmoidWeightedLU:2017}
Elfwing Stefan, Uchibe Eiji, and Doya Kenji.
\newblock Sigmoid-weighted linear units for neural network function
  approximation in reinforcement learning.
\newblock \emph{Neural networks : the official journal of the International
  Neural Network Society}, 107:\penalty0 3--11, 2017.

\bibitem[Stefan and Helmut(1998)]{Leopoldseder:ApproximationOD:1998}
Leopoldseder Stefan and Pottmann Helmut.
\newblock Approximation of developable surfaces with cone spline surfaces.
\newblock \emph{Computer-Aided Design}, 30:\penalty0 571--582, 1998.

\bibitem[Stein et~al.(2018)Stein, Grinspun, and
  Crane]{Odedstein:DevelopabilityTriangleMeshes:2018}
Oded Stein, Eitan Grinspun, and Keenan Crane.
\newblock Developability of triangle meshes.
\newblock \emph{ACM Transaction on Graphics (TOG)}, 37, 2018.

\bibitem[Tae-Hyun et~al.(2016{\natexlab{a}})Tae-Hyun, Yu-Wing, Jean-Charles,
  Hyeongwoo, and In~So]{Oh:PartialSum:2016}
Oh Tae-Hyun, Tai Yu-Wing, Bazin Jean-Charles, Kim Hyeongwoo, and Kweon In~So.
\newblock Partial sum minimization of singular values in robust pca: Algorithm
  and applications.
\newblock \emph{IEEE Transactions on Pattern Analysis and Machine
  Intelligence}, 38\penalty0 (4):\penalty0 744--758, 2016{\natexlab{a}}.

\bibitem[Tae-Hyun et~al.(2016{\natexlab{b}})Tae-Hyun, Yu-Wing, Jean-Charles,
  Hyeongwoo, and In~So]{Oh:Partialnuclearnorm:2016}
Oh Tae-Hyun, Tai Yu-Wing, Bazin Jean-Charles, Kim Hyeongwoo, and Kweon In~So.
\newblock Partial sum minimization of singular values in robust pca: Algorithm
  and applications.
\newblock \emph{IEEE Transactions on Pattern Analysis and Machine
  Intelligence}, 38\penalty0 (4):\penalty0 744--758, 2016{\natexlab{b}}.

\bibitem[Verhoeven et~al.(2022)Verhoeven, Vaxman, Hoffmann, and
  Sorkine-Hornung]{Verhoeven:Dev2PQ:2022}
Floor Verhoeven, Amir Vaxman, Tim Hoffmann, and Olga Sorkine-Hornung.
\newblock Dev2pq: Planar quadrilateral strip remeshing of developable surfaces.
\newblock \emph{ACM Transactions on Graphics (TOG)}, 41\penalty0 (3):\penalty0
  29:1–18, 2022.

\bibitem[Vincent et~al.(2020)Vincent, Julien~N.P., Alexander~W., David~B., and
  Gordon]{sitzmann:siren:2019}
Sitzmann Vincent, Martel Julien~N.P., Bergman Alexander~W., Lindell David~B.,
  and Wetzstein Gordon.
\newblock Implicit neural representations with periodic activation functions.
\newblock In \emph{Proc. NeurIPS}, 2020.

\bibitem[William~E. and Harvey~E.(1987)]{Lorensen:marchingcubes:1987}
Lorensen William~E. and Cline Harvey~E.
\newblock Marching cubes: A high resolution 3d surface construction algorithm.
\newblock In \emph{Proceedings of the 14th Annual Conference on Computer
  Graphics and Interactive Techniques}, page 163–169. Association for
  Computing Machinery, 1987.

\bibitem[Yoshua et~al.(2013)Yoshua, Nicholas, and
  Aaron]{Bengio:EstimatingOP:2013}
Bengio Yoshua, L{\'{e}}onard Nicholas, and C.~Courville Aaron.
\newblock Estimating or propagating gradients through stochastic neurons for
  conditional computation.
\newblock \emph{CoRR}, abs/1308.3432, 2013.

\bibitem[Zhiqin and Hao(2019)]{Chen:LearningImplicitFields:2019}
Chen Zhiqin and Zhang Hao.
\newblock Learning implicit fields for generative shape modeling.
\newblock In \emph{2019 IEEE/CVF Conference on Computer Vision and Pattern
  Recognition (CVPR)}, pages 5932--5941. IEEE Computer Society, 2019.

\end{thebibliography}
